\documentclass[11pt]{article}

\usepackage[T1]{fontenc}
\usepackage[utf8]{inputenc}
\usepackage[margin=1in]{geometry}
\usepackage{amsmath,amssymb}
\usepackage{booktabs}
\usepackage{array}
\usepackage{float}
\usepackage{longtable}
\usepackage[section]{placeins}
\usepackage{graphicx}
\usepackage{xcolor}
\usepackage[round]{natbib}
\usepackage[hidelinks]{hyperref}
\usepackage{caption}
\usepackage{pgfplots}
\pgfplotsset{compat=1.18}
\definecolor{c47m}{HTML}{2A78D6}
\definecolor{c107m}{HTML}{EB6834}
\definecolor{c349m}{HTML}{1BAF7A}
\definecolor{c16m}{HTML}{EDA100}
\definecolor{c204m}{HTML}{E87BA4}
\definecolor{c818m}{HTML}{6250D6}
\definecolor{c550m}{HTML}{8A8A8A}

\newcommand{\val}[1]{\ifcsname v@#1\endcsname\csname v@#1\endcsname\else\textcolor{red}{[#1]}\fi}
\expandafter\def\csname v@bdiff-range\endcsname{0.03 to 0.05}
\expandafter\def\csname v@bmult-107M-full\endcsname{$\times$1.21}
\expandafter\def\csname v@bmult-16M-double\endcsname{$\times$1.33}
\expandafter\def\csname v@bmult-16M-full\endcsname{$\times$1.34}
\expandafter\def\csname v@bmult-47M-double\endcsname{$\times$1.28}
\expandafter\def\csname v@bmult-47M-full\endcsname{$\times$1.30}
\expandafter\def\csname v@bmult-47M-half\endcsname{$\times$1.34}
\expandafter\def\csname v@bseeds\endcsname{seed 1337, and at 16M also seed 2}
\expandafter\def\csname v@eqslow-0.275\endcsname{5 to 9}
\expandafter\def\csname v@eqspeed-16M\endcsname{11}
\expandafter\def\csname v@eqspeed-349M\endcsname{3}
\expandafter\def\csname v@eqspeed-range\endcsname{3 to 11}
\expandafter\def\csname v@exch-107M-chorus1.09\endcsname{0.19}
\expandafter\def\csname v@exch-16M-chorus1.09\endcsname{0.31}
\expandafter\def\csname v@exch-204M-chorus1.09\endcsname{0.17}
\expandafter\def\csname v@exch-349M-chorus1.09\endcsname{0.14}
\expandafter\def\csname v@exch-47M-chorus1.09\endcsname{0.27}
\expandafter\def\csname v@exch-47M-chorusmax\endcsname{0.03}
\expandafter\def\csname v@fit-A\endcsname{2.2070}
\expandafter\def\csname v@fit-E\endcsname{2.4526}
\expandafter\def\csname v@fit-alpha\endcsname{0.254}
\expandafter\def\csname v@fixedsame-range\endcsname{30 to 38}
\expandafter\def\csname v@fixedshare-107M-chorus1.09\endcsname{12}
\expandafter\def\csname v@fixedshare-16M-chorus1.09\endcsname{25}
\expandafter\def\csname v@fixedshare-204M-chorus1.09\endcsname{10}
\expandafter\def\csname v@fixedshare-349M-chorus1.09\endcsname{9}
\expandafter\def\csname v@fixedshare-47M-chorus0.275\endcsname{64}
\expandafter\def\csname v@fixedshare-47M-chorus1.09\endcsname{16}
\expandafter\def\csname v@flopsadd-range\endcsname{7}
\expandafter\def\csname v@flopseq-range\endcsname{8 to 19}
\expandafter\def\csname v@gain-47M-bve-best\endcsname{0.012}
\expandafter\def\csname v@gain-47M-bve-lr1\endcsname{0.038}
\expandafter\def\csname v@gain-47M-chorus1.09-best\endcsname{0.049}
\expandafter\def\csname v@gain-47M-chorus1.09-lr0.5\endcsname{0.092}
\expandafter\def\csname v@gain-47M-chorus1.09-lr1\endcsname{0.077}
\expandafter\def\csname v@gain-47M-chorus1.09-lr32\endcsname{0.037}
\expandafter\def\csname v@gaincut-range\endcsname{34 to 54}
\expandafter\def\csname v@halfbranch-16M\endcsname{3.6271}
\expandafter\def\csname v@halffresh-16M\endcsname{3.6248}
\expandafter\def\csname v@loss-107M-b-best\endcsname{3.1257}
\expandafter\def\csname v@loss-47M-chorusmax-best\endcsname{3.1791}
\expandafter\def\csname v@loss-550M-b-best\endcsname{2.8976}
\expandafter\def\csname v@mult-107M-chorus1.09-best\endcsname{1.20}
\expandafter\def\csname v@mult-16M-chorus1.09-best\endcsname{1.34}
\expandafter\def\csname v@mult-204M-chorus1.09-best\endcsname{1.18}
\expandafter\def\csname v@mult-349M-chorus1.09-best\endcsname{1.16}
\expandafter\def\csname v@mult-47M-chorus1.09-best\endcsname{1.29}
\expandafter\def\csname v@multfit-107M-chorus1.09\endcsname{1.19}
\expandafter\def\csname v@multfit-16M-chorus1.09\endcsname{1.30}
\expandafter\def\csname v@multfit-204M-chorus1.09\endcsname{1.18}
\expandafter\def\csname v@multfit-349M-chorus1.09\endcsname{1.15}
\expandafter\def\csname v@multfit-47M-chorus1.09\endcsname{1.27}
\expandafter\def\csname v@multne-107M-chorus1.09\endcsname{1.21}
\expandafter\def\csname v@multne-16M-chorus1.09\endcsname{1.38}
\expandafter\def\csname v@multne-204M-chorus1.09\endcsname{1.19}
\expandafter\def\csname v@multne-349M-chorus1.09\endcsname{1.16}
\expandafter\def\csname v@multne-47M-chorus1.09\endcsname{1.31}
\expandafter\def\csname v@oggain-107M-13gramnone\endcsname{0.022}
\expandafter\def\csname v@oggain-107M-50gramnone\endcsname{0.026}
\expandafter\def\csname v@oggain-107M-all\endcsname{0.029}
\expandafter\def\csname v@oggain-16M-13gramnone\endcsname{0.063}
\expandafter\def\csname v@oggain-16M-50gramnone\endcsname{0.068}
\expandafter\def\csname v@oggain-16M-all\endcsname{0.070}
\expandafter\def\csname v@oggain-204M-13gramnone\endcsname{0.015}
\expandafter\def\csname v@oggain-204M-50gramnone\endcsname{0.019}
\expandafter\def\csname v@oggain-204M-all\endcsname{0.023}
\expandafter\def\csname v@oggain-349M-13gramnone\endcsname{0.011}
\expandafter\def\csname v@oggain-349M-50gramnone\endcsname{0.014}
\expandafter\def\csname v@oggain-349M-all\endcsname{0.017}
\expandafter\def\csname v@oggain-47M-13gramnone\endcsname{0.043}
\expandafter\def\csname v@oggain-47M-50gramnone\endcsname{0.046}
\expandafter\def\csname v@oggain-47M-all\endcsname{0.049}
\expandafter\def\csname v@ogmult-107M\endcsname{1.20}
\expandafter\def\csname v@ogmult-107M-13gramnone\endcsname{1.16}
\expandafter\def\csname v@ogmult-107M-50gramnone\endcsname{1.18}
\expandafter\def\csname v@ogmult-16M\endcsname{1.34}
\expandafter\def\csname v@ogmult-16M-13gramnone\endcsname{1.31}
\expandafter\def\csname v@ogmult-16M-50gramnone\endcsname{1.33}
\expandafter\def\csname v@ogmult-204M\endcsname{1.18}
\expandafter\def\csname v@ogmult-204M-13gramnone\endcsname{1.12}
\expandafter\def\csname v@ogmult-204M-50gramnone\endcsname{1.16}
\expandafter\def\csname v@ogmult-349M\endcsname{1.16}
\expandafter\def\csname v@ogmult-349M-13gramnone\endcsname{1.10}
\expandafter\def\csname v@ogmult-349M-50gramnone\endcsname{1.13}
\expandafter\def\csname v@ogmult-47M\endcsname{1.29}
\expandafter\def\csname v@ogmult-47M-13gramnone\endcsname{1.26}
\expandafter\def\csname v@ogmult-47M-50gramnone\endcsname{1.28}
\expandafter\def\csname v@ogmultci-107M\endcsname{1.199 to 1.207}
\expandafter\def\csname v@ogmultci-16M\endcsname{1.336 to 1.342}
\expandafter\def\csname v@ogmultci-204M\endcsname{1.179 to 1.188}
\expandafter\def\csname v@ogmultci-349M\endcsname{1.153 to 1.163}
\expandafter\def\csname v@ogmultci-47M\endcsname{1.290 to 1.296}
\expandafter\def\csname v@ogwin-13gramnone\endcsname{1{,}390}
\expandafter\def\csname v@ogwin-50gramnone\endcsname{6{,}433}
\expandafter\def\csname v@ogwin-all\endcsname{8{,}192}
\expandafter\def\csname v@ov13-checked\endcsname{200 of 200}
\expandafter\def\csname v@ov13-over50\endcsname{269}
\expandafter\def\csname v@ov13-share\endcsname{11.0}
\expandafter\def\csname v@ov50-any\endcsname{1{,}759}
\expandafter\def\csname v@ov50-checked\endcsname{200 of 200}
\expandafter\def\csname v@ov50-share\endcsname{1.0}
\expandafter\def\csname v@pred-550M-curve\endcsname{2.8982}
\expandafter\def\csname v@pred-550M-inputs\endcsname{3.5489, 3.2831, 3.1257, 3.0232 and 2.9535}
\expandafter\def\csname v@pred-550M-line\endcsname{2.8941}
\expandafter\def\csname v@quant-bits2\endcsname{0.186 to 0.235}
\expandafter\def\csname v@quant-bits3\endcsname{0.012 to 0.014}
\expandafter\def\csname v@quant-bits4\endcsname{0.002}
\expandafter\def\csname v@quant-bits8\endcsname{0.000}
\expandafter\def\csname v@quantkept-bits4\endcsname{97 to 98}
\expandafter\def\csname v@quantquiz-bits4\endcsname{1}
\expandafter\def\csname v@quiz-107M-b-best\endcsname{0.161}
\expandafter\def\csname v@quiz-107M-chorus1.09-best\endcsname{0.211}
\expandafter\def\csname v@quiz-107M-chorus1.09-lr2\endcsname{0.169}
\expandafter\def\csname v@quiz-107M-chorus1.09-lr4\endcsname{0.185}
\expandafter\def\csname v@quiz-107M-chorus1.09-lr8\endcsname{0.211}
\expandafter\def\csname v@quiz-16M-b-best\endcsname{0.076}
\expandafter\def\csname v@quiz-16M-chorus1.09-best\endcsname{0.089}
\expandafter\def\csname v@quiz-204M-b-best\endcsname{0.230}
\expandafter\def\csname v@quiz-204M-chorus1.09-best\endcsname{0.218}
\expandafter\def\csname v@quiz-349M-b-best\endcsname{0.233}
\expandafter\def\csname v@quiz-349M-chorus1.09-best\endcsname{0.240}
\expandafter\def\csname v@quiz-47M-b-best\endcsname{0.123}
\expandafter\def\csname v@quiz-47M-chorus1.09-best\endcsname{0.146}
\expandafter\def\csname v@quizeq-107M\endcsname{0.180}
\expandafter\def\csname v@quizeq-16M\endcsname{0.089}
\expandafter\def\csname v@quizeq-204M\endcsname{0.231}
\expandafter\def\csname v@quizeq-349M\endcsname{0.245}
\expandafter\def\csname v@quizeq-47M\endcsname{0.135}
\expandafter\def\csname v@quizzero-107M-chorus1.09-best\endcsname{0.045}
\expandafter\def\csname v@quizzero-16M-chorus1.09-best\endcsname{0.054}
\expandafter\def\csname v@quizzero-204M-chorus1.09-best\endcsname{0.054}
\expandafter\def\csname v@quizzero-349M-chorus1.09-best\endcsname{0.067}
\expandafter\def\csname v@quizzero-47M-chorus1.09-best\endcsname{0.060}
\expandafter\def\csname v@rows-47M-best\endcsname{0.038}
\expandafter\def\csname v@rows-47M-lr1\endcsname{0.039}
\expandafter\def\csname v@runs\endcsname{115}
\expandafter\def\csname v@seedcells\endcsname{23}
\expandafter\def\csname v@seedsizes\endcsname{16M, 47M, 107M, 204M and 349M}
\expandafter\def\csname v@seedspread-loss\endcsname{0.0002 to 0.0088}
\expandafter\def\csname v@seedspread-quiz\endcsname{0.001 to 0.033}
\expandafter\def\csname v@side-ember\endcsname{3.2660}
\expandafter\def\csname v@side-ember-lr0p0023\endcsname{3.2601}
\expandafter\def\csname v@side-ember-lr0p007\endcsname{3.2590}
\expandafter\def\csname v@side-ember-lr0p023\endcsname{3.2604}
\expandafter\def\csname v@side-ember-lr0p073\endcsname{3.2626}
\expandafter\def\csname v@side-ember-lr0p73\endcsname{3.2685}
\expandafter\def\csname v@side-emberlazy\endcsname{3.2672}
\expandafter\def\csname v@side-emberlazy-lr0p023\endcsname{3.2585}
\expandafter\def\csname v@side-tablerate0p33\endcsname{3.2539}
\expandafter\def\csname v@side-tablerate0p33-lr8\endcsname{3.2290}
\expandafter\def\csname v@side-tablerate1\endcsname{3.2531}
\expandafter\def\csname v@side-tablerate1-lr8\endcsname{3.2332}
\expandafter\def\csname v@side-tablerate3\endcsname{3.2498}
\expandafter\def\csname v@side-tablerate3-lr8\endcsname{3.2498}
\expandafter\def\csname v@side-tablerate9\endcsname{3.2514}
\expandafter\def\csname v@signa-107M\endcsname{58}
\expandafter\def\csname v@signa-16M\endcsname{25}
\expandafter\def\csname v@signa-204M\endcsname{40}
\expandafter\def\csname v@signa-349M\endcsname{47}
\expandafter\def\csname v@signa-47M\endcsname{32}
\expandafter\def\csname v@signb-107M\endcsname{12}
\expandafter\def\csname v@signb-16M\endcsname{17}
\expandafter\def\csname v@signb-204M\endcsname{48}
\expandafter\def\csname v@signb-349M\endcsname{42}
\expandafter\def\csname v@signb-47M\endcsname{18}
\expandafter\def\csname v@signp-107M\endcsname{$<0.001$}
\expandafter\def\csname v@signp-16M\endcsname{$0.28$}
\expandafter\def\csname v@signp-204M\endcsname{$0.46$}
\expandafter\def\csname v@signp-349M\endcsname{$0.67$}
\expandafter\def\csname v@signp-47M\endcsname{$0.06$}
\expandafter\def\csname v@signseed-107M\endcsname{20 against 43, $p=0.005$}
\expandafter\def\csname v@slowdown\endcsname{9 to 16}
\expandafter\def\csname v@tokspread\endcsname{3}
\expandafter\def\csname v@trdiff-range\endcsname{0.004 to 0.009}
\expandafter\def\csname v@treq-107M-tr0p33\endcsname{136M}
\expandafter\def\csname v@treq-16M-tr0p04\endcsname{21.3M}
\expandafter\def\csname v@treq-16M-tr0p11\endcsname{21.4M}
\expandafter\def\csname v@treq-16M-tr0p33\endcsname{21.3M}
\expandafter\def\csname v@treq-204M-tr0p33\endcsname{250M}
\expandafter\def\csname v@treq-349M-tr0p33\endcsname{418M}
\expandafter\def\csname v@treq-47M-tr0p11\endcsname{62.0M}
\expandafter\def\csname v@treq-47M-tr0p33\endcsname{62.1M}
\expandafter\def\csname v@treqdef-107M-tr0p33\endcsname{129M}
\expandafter\def\csname v@treqdef-16M-tr0p04\endcsname{21.0M}
\expandafter\def\csname v@treqdef-16M-tr0p11\endcsname{21.0M}
\expandafter\def\csname v@treqdef-16M-tr0p33\endcsname{21.0M}
\expandafter\def\csname v@treqdef-204M-tr0p33\endcsname{242M}
\expandafter\def\csname v@treqdef-349M-tr0p33\endcsname{404M}
\expandafter\def\csname v@treqdef-47M-tr0p11\endcsname{60.9M}
\expandafter\def\csname v@treqdef-47M-tr0p33\endcsname{60.9M}
\expandafter\def\csname v@trloss-107M-tr0p33\endcsname{3.0875}
\expandafter\def\csname v@trloss-16M-tr0p04\endcsname{3.4748}
\expandafter\def\csname v@trloss-16M-tr0p11\endcsname{3.4743}
\expandafter\def\csname v@trloss-16M-tr0p33\endcsname{3.4755}
\expandafter\def\csname v@trloss-204M-tr0p33\endcsname{2.9962}
\expandafter\def\csname v@trloss-349M-tr0p33\endcsname{2.9298}
\expandafter\def\csname v@trloss-47M-tr0p11\endcsname{3.2303}
\expandafter\def\csname v@trloss-47M-tr0p33\endcsname{3.2300}
\expandafter\def\csname v@trlossdef-107M-tr0p33\endcsname{3.0966}
\expandafter\def\csname v@trlossdef-16M-tr0p04\endcsname{3.4791}
\expandafter\def\csname v@trlossdef-16M-tr0p11\endcsname{3.4791}
\expandafter\def\csname v@trlossdef-16M-tr0p33\endcsname{3.4791}
\expandafter\def\csname v@trlossdef-204M-tr0p33\endcsname{3.0007}
\expandafter\def\csname v@trlossdef-349M-tr0p33\endcsname{2.9339}
\expandafter\def\csname v@trlossdef-47M-tr0p11\endcsname{3.2337}
\expandafter\def\csname v@trlossdef-47M-tr0p33\endcsname{3.2337}
\expandafter\def\csname v@trmult-107M-tr0p33\endcsname{1.27}
\expandafter\def\csname v@trmult-16M-tr0p04\endcsname{1.36}
\expandafter\def\csname v@trmult-16M-tr0p11\endcsname{1.36}
\expandafter\def\csname v@trmult-16M-tr0p33\endcsname{1.36}
\expandafter\def\csname v@trmult-204M-tr0p33\endcsname{1.22}
\expandafter\def\csname v@trmult-349M-tr0p33\endcsname{1.20}
\expandafter\def\csname v@trmult-47M-tr0p11\endcsname{1.32}
\expandafter\def\csname v@trmult-47M-tr0p33\endcsname{1.32}
\expandafter\def\csname v@trmultdef-107M-tr0p33\endcsname{1.20}
\expandafter\def\csname v@trmultdef-16M-tr0p04\endcsname{1.34}
\expandafter\def\csname v@trmultdef-16M-tr0p11\endcsname{1.34}
\expandafter\def\csname v@trmultdef-16M-tr0p33\endcsname{1.34}
\expandafter\def\csname v@trmultdef-204M-tr0p33\endcsname{1.18}
\expandafter\def\csname v@trmultdef-349M-tr0p33\endcsname{1.16}
\expandafter\def\csname v@trmultdef-47M-tr0p11\endcsname{1.29}
\expandafter\def\csname v@trmultdef-47M-tr0p33\endcsname{1.29}
\expandafter\def\csname v@trmultup-range\endcsname{0.02 to 0.07}
\expandafter\def\csname v@trseeds-107M-tr0p33\endcsname{2}
\expandafter\def\csname v@trseeds-16M-tr0p04\endcsname{5}
\expandafter\def\csname v@trseeds-16M-tr0p11\endcsname{5}
\expandafter\def\csname v@trseeds-16M-tr0p33\endcsname{5}
\expandafter\def\csname v@trseeds-204M-tr0p33\endcsname{2}
\expandafter\def\csname v@trseeds-349M-tr0p33\endcsname{2}
\expandafter\def\csname v@trseeds-47M-tr0p11\endcsname{3}
\expandafter\def\csname v@trseeds-47M-tr0p33\endcsname{3}
\expandafter\def\csname v@vbudget-32k-1x\endcsname{0.025}
\expandafter\def\csname v@vbudget-32k-2x\endcsname{0.023}
\expandafter\def\csname v@vbudget-32k-4x\endcsname{0.022}
\expandafter\def\csname v@vbudget-64k-1x\endcsname{0.018}
\expandafter\def\csname v@vbudget-64k-2x\endcsname{0.017}
\expandafter\def\csname v@vbudget-8k-1x\endcsname{0.034}
\expandafter\def\csname v@vbudget-8k-2x\endcsname{0.033}
\expandafter\def\csname v@vgain-47M-64k\endcsname{0.018}
\expandafter\def\csname v@vgain-47M-8k\endcsname{0.033}
\expandafter\def\csname v@vgain-seeddiff\endcsname{0.002}
\expandafter\def\csname v@vrows-64k-many\endcsname{0.024}
\expandafter\def\csname v@vrows-8k-few\endcsname{0.021}

\title{Bigger Model or Bigger Table?\\
How Much Backbone a Chorus N-gram Module Replaces}
\author{Szymon Kocur\\
\small Independent Researcher, Warsaw, Poland\\
\small \href{mailto:me@szymonkocur.com}{\texttt{me@szymonkocur.com}}\quad
ORCID: \href{https://orcid.org/0009-0005-3372-5133}{0009-0005-3372-5133}}
\date{November 2026}


\begin{document}
\maketitle

\begin{center}
\begin{tikzpicture}
\begin{axis}[name=left, width=0.52\linewidth, height=0.42\linewidth, xmode=log,
  xlabel={Backbone parameters (millions)}, ylabel={Held-out loss (nats)},
  xtick={16,47,107,204,349,550}, xticklabels={16,47,107,204,349,550}, xmin=12, xmax=700,
  legend pos=north east, legend cell align=left, legend style={font=\scriptsize, draw=none, fill=none},
  tick label style={font=\scriptsize}, label style={font=\footnotesize},
  grid=major, grid style={gray!25}, title={\footnotesize (a) reading the equivalent size}]
\addplot[black!75, thick, solid, mark=*] coordinates {(15.7,3.5496) (47.1,3.2831) (106.9,3.1257) (204.4,3.0236) (349.1,2.9511) (550.4,2.8976)};
\addlegendentry{plain model}
\addplot[c16m, only marks, mark=square*, mark size=2.5pt, forget plot] coordinates {(15.7,3.4791)};
\draw[->, thick, c16m, solid] (axis cs:15.7,3.4791) -- (axis cs:21.0,3.4791);
\addplot[c47m, only marks, mark=square*, mark size=2.5pt, forget plot] coordinates {(47.1,3.2337)};
\draw[->, thick, c47m, solid] (axis cs:47.1,3.2337) -- (axis cs:60.9,3.2337);
\addplot[c107m, only marks, mark=square*, mark size=2.5pt, forget plot] coordinates {(106.9,3.0966)};
\draw[->, thick, c107m, solid] (axis cs:106.9,3.0966) -- (axis cs:128.5,3.0966);
\addplot[c204m, only marks, mark=square*, mark size=2.5pt, forget plot] coordinates {(204.4,3.0007)};
\draw[->, thick, c204m, solid] (axis cs:204.4,3.0007) -- (axis cs:241.9,3.0007);
\addplot[c349m, only marks, mark=square*, mark size=2.5pt, forget plot] coordinates {(349.1,2.9339)};
\draw[->, thick, c349m, solid] (axis cs:349.1,2.9339) -- (axis cs:404.1,2.9339);
\addlegendimage{only marks, mark=square*, mark size=2.5pt, black!45}
\addlegendentry{Chorus, equal-size tables}
\addlegendimage{black!45, thick, ->}
\addlegendentry{to the size of equal loss}

\end{axis}
\begin{axis}[at={(left.east)}, anchor=west, xshift=0.1\linewidth, width=0.5\linewidth,
  height=0.42\linewidth, xmode=log, ymode=log,
  xlabel={Added parameters / backbone}, ylabel={Multiplier (log scale)},
  ytick={1,1.25,1.5,2,2.5}, yticklabels={1,1.25,1.5,2,2.5}, ymin=1, ymax=2.6,
  log basis x=2, xticklabel={$\pgfmathparse{2^\tick}\pgfmathprintnumber{\pgfmathresult}$},
  legend pos=north west, legend cell align=left, legend style={font=\scriptsize, draw=none, fill=none},
  tick label style={font=\scriptsize}, label style={font=\footnotesize},
  grid=major, grid style={gray!25}, title={\footnotesize (b) multiplier against added parameters}]
\addplot[c16m, thick, dotted, mark=oplus*, mark options={solid}] coordinates {(0.2235,1.0914) (0.8856,1.3381) (3.5342,1.6489) (17.6599,1.9669)};
\addlegendentry{16M}
\addplot[c47m, thick, solid, mark=square*, mark options={solid}] coordinates {(0.2479,1.0688) (0.9826,1.2932) (3.9214,1.4757) (19.5949,1.7178)};
\addlegendentry{47M}
\addplot[c107m, thick, dashed, mark=triangle*, mark options={solid}] coordinates {(0.2751,1.0190) (1.0903,1.2028) (4.3511,1.3835) (9.5842,1.4813)};
\addlegendentry{107M}
\addplot[c204m, thick, dashdotted, mark=diamond*, mark options={solid}] coordinates {(0.3052,1.0408) (1.2098,1.1839) (4.8278,1.3336) (5.5604,1.3395)};
\addlegendentry{204M}
\addplot[c349m, thick, densely dashed, mark=pentagon*, mark options={solid}] coordinates {(1.3423,1.1577)};
\addlegendentry{349M}

\end{axis}
\end{tikzpicture}

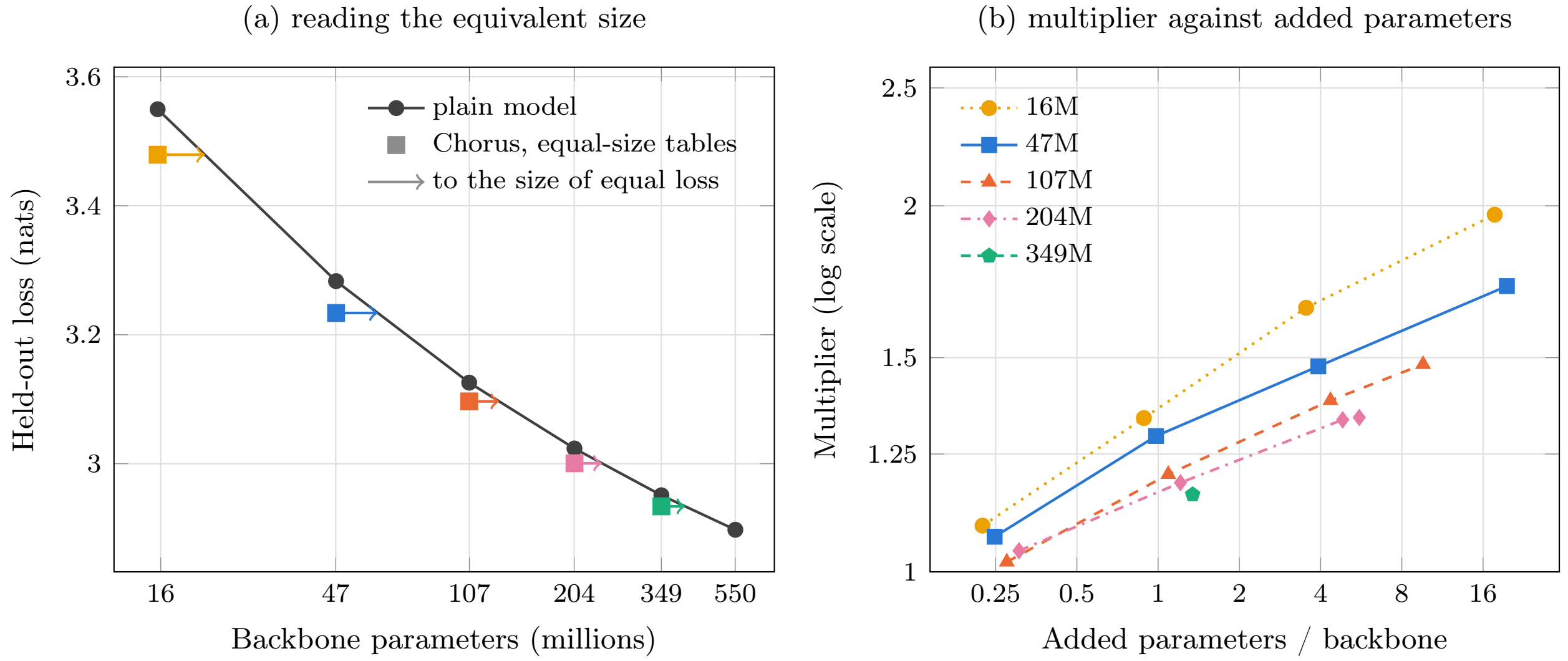
\captionof{figure}{(a) Held-out loss against backbone size. Circles: plain models. Squares: the same
models with Chorus, hashed N-gram tables added to the transformer (the backbone), adding about as many
parameters as the backbone. Every model is at its
best measured backbone learning rate (Section~\ref{sec:training}). Each arrow ends at the equivalent size, the size a plain transformer would need for the
same loss. (b) The multiplier, equivalent size divided by backbone size, against all parameters Chorus
adds, its token table, gates and projections included, divided by the backbone's. Other table sizes are at the rates of Table~\ref{tab:worth}. Sizes are shifted slightly sideways so that their points do not cover
each other; Table~\ref{tab:worth} gives the exact values.}
\label{fig:teaser}
\end{center}

\begin{abstract}
We ask how much of a transformer's backbone the Chorus N-gram module replaces. Chorus adds hashed
tables of token bigrams and trigrams to its input and attention. We train models of 16M to 349M parameters,
and a plain one of 550M, on 1.34 billion tokens of Polish printed before 1918 and find the size of a plain
model with the same held-out loss. A separate learning-rate search for each kind of model cuts Chorus's gain
by \val{gaincut-range}\%. With tables about as large as the backbone, Chorus then matches a plain model
\val{mult-16M-chorus1.09-best} times its size at 16M and \val{mult-349M-chorus1.09-best} to
\val{mult-107M-chorus1.09-best} times from 107M to 349M. With these tables, each added parameter
replaces at most about a third of a backbone parameter, and Chorus trains slower than a plain model of its
size and \val{eqspeed-range}\% faster than one of the same loss. With vocabularies of up to 64,000 tokens at
one shared learning rate, Chorus gains less from 16M to 107M. The results rest on one corpus and a fifth of
an epoch.
\end{abstract}

\section{Introduction}
\label{sec:intro}

Lookup memories give a language model more parameters with little added computation per token.
Engram \citep{cheng2026engram} reads hashed N-gram embeddings into the residual stream, and Chorus
\citep{chung2026chorus} shares one set of N-gram tables between the input and the attention values of
many layers. Chung and Du vary Chorus's tables from 20M to 9.6B parameters on a backbone of 21M and train one table
size each on backbones of 17M and 133M.
Someone choosing a model needs to know how many backbone parameters a table of a given size replaces,
and whether that number changes with the size of the backbone.

We measure this number for Chorus at five backbone sizes, from 16M to 349M parameters, on the
Wieszcz-XIX corpus of Polish printed before 1918 \citep{kocur2026wieszcz}. We call the transformer
that Chorus is added to the backbone, and the same transformer trained without Chorus the plain model.
For each Chorus model we read off the size a plain model needs for the same held-out loss, its
equivalent size. A learning rate shared by both kinds of model can favour one of them, so we search the
rate at every size for each kind separately. Section~\ref{sec:results} also reports other token
budgets, larger tables, the module's parts, vocabularies, a fact quiz and throughput.

\section{Related work}
\label{sec:related}

\paragraph{N-gram and token tables.} Several recent modules add parameters that are looked up by
token or by a hash of the last few tokens. Engram \citep{cheng2026engram} reads such embeddings into a mixture-of-experts model. The Over-Tokenized
Transformer \citep{huang2025overtokenized} grows the input vocabulary with them and reports that its 400M
model with the largest vocabulary matches the training loss of a 1B model without it. The count tables
of COUNT, held in host memory \citep{magnusson2026count}, keep statistics of the training data outside
the network. Appendix~\ref{sec:further} lists further work of this kind.

\paragraph{Equivalent sizes of added parameters.} Scaling studies relate loss to model size and data
\citep{kaplan2020scaling, hoffmann2022chinchilla}, and untuned learning rates distort such
comparisons \citep{porian2024resolving}. N-gram tables added to a pretrained 0.6B model brought no gain
at the backbone's rate and were trained at fifty times that rate \citep{hu2026augmenting}. The value of
added parameters has been given as the size of an equivalent dense model for routed models
\citep{clark2022unified}, and memory layers outperform dense models with more than twice their
computation \citep{berges2025memory}. JTok \citep{yang2026jtok} reports the computation a
mixture-of-experts model saves with token-indexed tables. Chorus extends a finding made by an AI agent on
rStation \citeyearpar{rstation2026gp27}, where agents also report dense-equivalent multipliers of hashed
N-gram input tables of 151k rows, about $\times$1.3 at 26M and $\times$1.2 at 135M, with one fixed recipe
\citeyearpar{rstation2026gp10}. The Over-Tokenized Transformer and COUNT each give their gain
across model sizes as an equivalent size or as computation, with one training recipe for models with
and without tables. To our knowledge, as of November 2026, no published work measures the equivalent
size of a learned N-gram memory with the backbone's learning rate searched at every size for both
kinds of model.

\section{Setup}
\label{sec:setup}

\subsection{Data}

All models train on the training split of the Wieszcz-XIX corpus \citep{kocur2026wieszcz}, 6.69
billion tokens of Polish printed before 1918, under the corpus's byte-level BPE tokenizer with a
vocabulary of 8{,}000 tokens. We measure held-out loss on the corpus's held-out split of 56.4 million
tokens. Near-identical documents are removed first. The split is then drawn per document, and held-out
documents with half or more of their five-word sequences in the training split are moved to the
training side. A full run reads 1.34 billion tokens at random positions
of the training split, a fifth of an epoch. Appendix~\ref{sec:details} gives the exact counts.

\subsection{Backbones}

The backbone is a decoder-only transformer with rotary position embeddings, RMSNorm before each
sublayer and on its output, QK-normalisation, a SwiGLU feed-forward layer, grouped-query attention, a
context of 1{,}024 tokens and input embeddings tied to the output head. The six sizes follow one rule:
the width is 48 times the number of layers, the head dimension is 64, and there are a third as many
key/value heads as query heads (Table~\ref{tab:sizes}).

\begin{table}[H]
\centering
\small
\begin{tabular}{lrrrrr}
\toprule
Name & Parameters & Without embedding & Layers & Width & Heads (KV) \\
\midrule
16M  & 15{,}668{,}608  & 12{,}596{,}608  & 8  & 384  & 6 (2) \\
47M  & 47{,}105{,}088  & 42{,}497{,}088  & 12 & 576  & 9 (3) \\
107M & 106{,}859{,}264 & 100{,}715{,}264 & 16 & 768  & 12 (4) \\
204M & 204{,}368{,}320 & 196{,}688{,}320 & 20 & 960  & 15 (5) \\
349M & 349{,}069{,}440 & 339{,}853{,}440 & 24 & 1152 & 18 (6) \\
550M & 550{,}399{,}808 & 539{,}647{,}808 & 28 & 1344 & 21 (7) \\
\bottomrule
\end{tabular}
\caption{Backbone sizes. Parameters counts every parameter of the plain model, including the tied
embedding of $8{,}000 \times$ width. The size axis of every result uses this count, except where
Section~\ref{sec:worth} reads it without the embedding. 550M is trained as a plain model only.}
\label{tab:sizes}
\end{table}

\subsection{Chorus}
\label{sec:chorus}

We reimplement Chorus as described by \citet{chung2026chorus}, following their code. For each token, the module
looks up one row in each of four shared N-gram tables, two for bigrams and two for trigrams, and one
row in a token table. The key of an N-gram is its token ids read as one number in base 8{,}001, the
extra symbol marking a position before the start of the sequence. A table's row is that key modulo its
number of rows, which is $m$, $m+2$, $m+4$ or $m+6$ for a size $m$ set by the table budget. Each N-gram row has a quarter of the model's width;
joined, they are added to the token row of the full width. A sigmoid gate computed from the last three
token embeddings scales the result, and a causal depthwise convolution over four positions refines it.
The result is added once to the input embedding and, through a projection gated by each block's hidden
state, to the attention values of the even-numbered blocks. The tables start at zero, so an untrained module leaves the
backbone unchanged.

Our code departs from the authors' in two ways. The authors' backbones normalise the input embedding;
ours does not, and the module's output is added to the raw embedding. The N-gram tables and the
token table, together called the tables, are trained by SparseAdam at a rate of 0.0015 without weight
decay. The authors use their Ember optimiser, which keeps one second-moment value per table row and per
column. Appendix~\ref{sec:details} has 47M runs with
Ember at six rates.

\paragraph{Fixed part and N-gram rows.} Chorus adds two groups of parameters. The fixed part is the
token table, the gates, the convolution and the projections, and its size follows from the backbone's
width and depth (8.26 million parameters at 47M, 4.61 million of them in the token table). The N-gram
rows are the rest, and we choose their number. We also train the fixed part alone, as Chorus without
N-gram tables, and call it ``fixed part only''.

\paragraph{Added ratio.} We describe a Chorus model by its added ratio, all the parameters Chorus adds
divided by the backbone's. We train ratios of 0.275, 1.09 and 4.35, and for the largest tables 21.7 at 16M
and 47M, 9.58 at 107M and 5.01 at 204M. At 349M we train 1.09 only. Tables of ratio 1.09 add about as many parameters as the backbone has,
and we call them equal-size tables. Their fixed part is \val{fixedshare-16M-chorus1.09}\% of what
Chorus adds at 16M and \val{fixedshare-349M-chorus1.09}\% at 349M.

\subsection{Training}
\label{sec:training}

A full run takes 5{,}104 steps of 262{,}144 tokens (256 sequences of 1{,}024) with a warmup of 300
steps. The rate then stays at its peak and falls linearly to a tenth of it over the last 10\% of
steps, a short form of the cooldown studied by \citet{hagele2024scaling}. Runs at half and
twice the budget of Table~\ref{tab:budget} resume a full run from a step still at the peak rate (exceptions in its caption) and cool down over their own
last 10\% of steps. Muon \citep{jordan2024muon}
trains the two-dimensional weights of the blocks and of the module, AdamW \citep{loshchilov2019adamw}
the embedding, the normalisation weights and the convolution, and SparseAdam the tables.

\paragraph{Learning rate per size.} Every size starts from the same peak rates, 0.015 for Muon and
0.0003 for AdamW, which we call the base rate. We name each rate by its
Muon value, with AdamW in proportion: Muon 0.12 comes with AdamW 0.0024. We search the rate separately
for the plain model and for Chorus with equal-size tables, in steps of a factor of two
(Table~\ref{tab:lr}). From 107M upward we ran fewer rates, and then the rates next to the best one. The
550M plain model ran at 0.06 and 0.12 only. A model's best measured rate is the one with the lowest held-out loss,
averaged over seeds; of two rates within a thousandth of a nat, we take the one with more seeds. We
mark it $^\ddagger$ when it lies at the edge of the searched range.

The fixed part only and the other table sizes run at 0.12 at 16M and 47M and at 0.06 at 107M and 204M,
and half of them also at the base rate (Table~\ref{tab:worth}). Equal-size tables also run at the base rate at every size. The tables'
own rate stays at 0.0015 at every size, five times the base AdamW rate.

\subsection{Measurement}
\label{sec:measure}

\paragraph{Held-out loss.} We measure cross-entropy in nats per token on 8{,}192 non-overlapping
windows of 1{,}024 tokens spread evenly over the held-out split, in \texttt{float32}. Every model is
scored on the same windows, so differences between models are paired. Chorus's gain is the plain
model's loss minus the Chorus model's at the same size. Every configuration is trained with seed 1337,
and \val{seedcells} configurations at \val{seedsizes} also with further seeds (Table~\ref{tab:runs}).
Values are averaged over seeds, which in Table~\ref{tab:runs} differ by \val{seedspread-loss} nats.
We call this range the seed spread. Plain and Chorus models with one seed share the backbone's initial weights and training positions,
since the module has its own generator.

\paragraph{Knowledge.} Equal loss can hide different downstream behaviour \citep{liu2023same}, so
we also use a fact quiz of 497 facts in seven groups, asked as 755 questions in the
spelling of the period (Appendix~\ref{sec:quizitems}). One group has no candidate list and is left
out, which leaves 688 scored questions. In the other groups every answer of the group is a candidate, scored as the continuation of
the question by the summed log-probability of its tokens. A question is known when the right answer
scores highest, and the score is the share of the 688 scored questions known. We compare two models
with a sign test over the questions exactly one of them knows. We also run the quiz with the N-gram
tables set to zero and the token table kept. Between seeds the score differs by
\val{seedspread-quiz}.

\paragraph{Cost.} We measure training throughput in tokens per second on the card, as the median over
the run with its first tenth left out. Each Chorus model is compared with the plain model of its size on
the same card type, and with a plain model of the same loss, whose throughput we interpolate between
the measured sizes on logarithmic axes. We count training FLOPs per token after
\citet{kaplan2020scaling}: six per parameter in a matrix product, the tied embedding once as the output
head, and six times the layers, the context and the width for attention. Chorus's table reads are
lookups, and its gates, convolution and projections count as matrix products.

\subsection{Equivalent size}
\label{sec:analysis}

We plot the plain models' held-out loss against size on a logarithmic
axis and join neighbouring sizes by straight lines. A Chorus model's equivalent size is the size at
which this line reaches the Chorus model's loss. Its multiplier is the equivalent size divided by the
backbone's size; a multiplier of $\times$1.5 means that a plain model needs one and a half times as many
parameters for the same loss. The backbone parameters replaced per added parameter are the equivalent
size minus the backbone's size, divided by the parameters Chorus adds. Beyond the largest plain model
we extend the last line and mark the value $^\dagger$.

\section{Results}
\label{sec:results}

\subsection{Learning rate per size}
\label{sec:lr}

\begin{table}[tbp]
\centering
\small
\begin{tabular}{llrrrrrrr}
\toprule
 & & \multicolumn{7}{c}{Peak Muon learning rate (AdamW in proportion)} \\
\cmidrule(l){3-9}
Size & Model & 0.0075 & 0.015 & 0.03 & 0.06 & 0.12 & 0.24 & 0.48 \\
\midrule
16M & plain & 3.6606 & 3.6071 & 3.5820 & 3.5685 & 3.5573 & \textbf{3.5496} & 3.5512 \\
16M & Chorus 1.09 & 3.5280 & 3.4997 & 3.4892 & 3.4803 & 3.4792 & \textbf{3.4791} & 3.4869 \\
16M & \emph{gain} & \emph{0.133} & \emph{0.107} & \emph{0.093} & \emph{0.088} & \emph{0.078} & \emph{0.071} & \emph{0.064} \\
\addlinespace
47M & plain & 3.3633 & 3.3301 & 3.3128 & 3.2946 & 3.2869 & \textbf{3.2831} & 3.2946 \\
47M & Chorus 1.09 & 3.2715 & 3.2531 & 3.2437 & 3.2368 & \textbf{3.2337} & 3.2410 & 3.2575 \\
47M & \emph{gain} & \emph{0.092} & \emph{0.077} & \emph{0.069} & \emph{0.058} & \emph{0.053} & \emph{0.042} & \emph{0.037} \\
\addlinespace
107M & plain & $\cdot$ & 3.1663 & 3.1489 & 3.1378 & 3.1270 & \textbf{3.1257} & 3.1463 \\
107M & Chorus 1.09 & $\cdot$ & 3.1036 & 3.1002 & 3.0969 & \textbf{3.0966} & 3.1061 & $\cdot$ \\
107M & \emph{gain} &  & \emph{0.063} & \emph{0.049} & \emph{0.041} & \emph{0.030} & \emph{0.020} &  \\
\addlinespace
204M & plain & $\cdot$ & 3.0522 & 3.0364 & 3.0290 & \textbf{3.0236} & 3.0286 & $\cdot$ \\
204M & Chorus 1.09 & $\cdot$ & 3.0109 & 3.0037 & \textbf{3.0007} & 3.0005 & $\cdot$ & $\cdot$ \\
204M & \emph{gain} &  & \emph{0.041} & \emph{0.033} & \emph{0.028} & \emph{0.023} &  &  \\
\addlinespace
349M & plain & $\cdot$ & 2.9763 & 2.9617 & 2.9543 & \textbf{2.9511} & 2.9596 & $\cdot$ \\
349M & Chorus 1.09 & $\cdot$ & 2.9434 & 2.9366 & \textbf{2.9339} & 2.9348 & $\cdot$ & $\cdot$ \\
349M & \emph{gain} &  & \emph{0.033} & \emph{0.025} & \emph{0.020} & \emph{0.016} &  &  \\
\addlinespace
550M & plain & $\cdot$ & $\cdot$ & $\cdot$ & \textbf{2.8976} & 2.8987 & $\cdot$ & $\cdot$ \\
\addlinespace
\bottomrule
\end{tabular}

\caption{Held-out loss in nats per token of plain models and of Chorus with equal-size tables at each
Muon learning rate, with Chorus's gain at each rate in italics. Bold: the best measured rate, the lowest loss after the tie rule of Section~\ref{sec:training}.
Chorus 1.09: Chorus with equal-size tables. $\cdot$: not run; larger models were run at fewer rates,
chosen from the best rates at the smaller sizes. Cells with more than one seed (Table~\ref{tab:runs})
give the average. Seeds of one configuration differ by
\val{seedspread-loss} nats.}
\label{tab:lr}
\end{table}

\paragraph{Best rates.} Both kinds of model are lowest well above the base rate (Table~\ref{tab:lr}). The
plain model is lowest at Muon 0.24 up to 107M and at 0.12 at 204M and 349M. The 550M model is lowest at
0.06, the lower of its two rates, and a lower loss at another rate would lower the 349M multiplier.

Chorus's best rate is 0.24 at 16M, 0.12 at 47M and 107M, and 0.06 at 204M and 349M. At every size
but 47M its loss at the next rate is within a thousandth of a nat.

\paragraph{Gain against rate.} At a rate shared by both kinds of model, Chorus's gain shrinks as the
rate rises, at every size and over the rates measured. At 47M it falls from \val{gain-47M-chorus1.09-lr0.5} nats at a Muon
rate of 0.0075 to \val{gain-47M-chorus1.09-lr32} at 0.48. With each model at its own best rate, the
gain is smaller than at the base rate at every size (Table~\ref{tab:worth}).

\subsection{Equivalent size}
\label{sec:worth}

Table~\ref{tab:worth} in Appendix~\ref{sec:worthall} gives, for each Chorus model, the size a plain model
needs for the same held-out loss. Figure~\ref{fig:teaser} plots its upper block.

\begin{figure}[H]
\centering
\begin{tikzpicture}
\begin{axis}[width=0.8\linewidth, height=0.36\linewidth, xmode=log,
  xlabel={Backbone parameters (millions)}, ylabel={Multiplier},
  xtick={16,47,107,204,349}, xticklabels={16,47,107,204,349}, xmin=12, xmax=450, ymin=1.1, ymax=1.6,
  ytick={1.1,1.2,1.3,1.4,1.5,1.6},
  legend pos=south west, legend cell align=left, legend style={font=\scriptsize, draw=none, fill=white, fill opacity=0.8, text opacity=1},
  tick label style={font=\scriptsize}, label style={font=\footnotesize}, grid=major, grid style={gray!25}]
\input{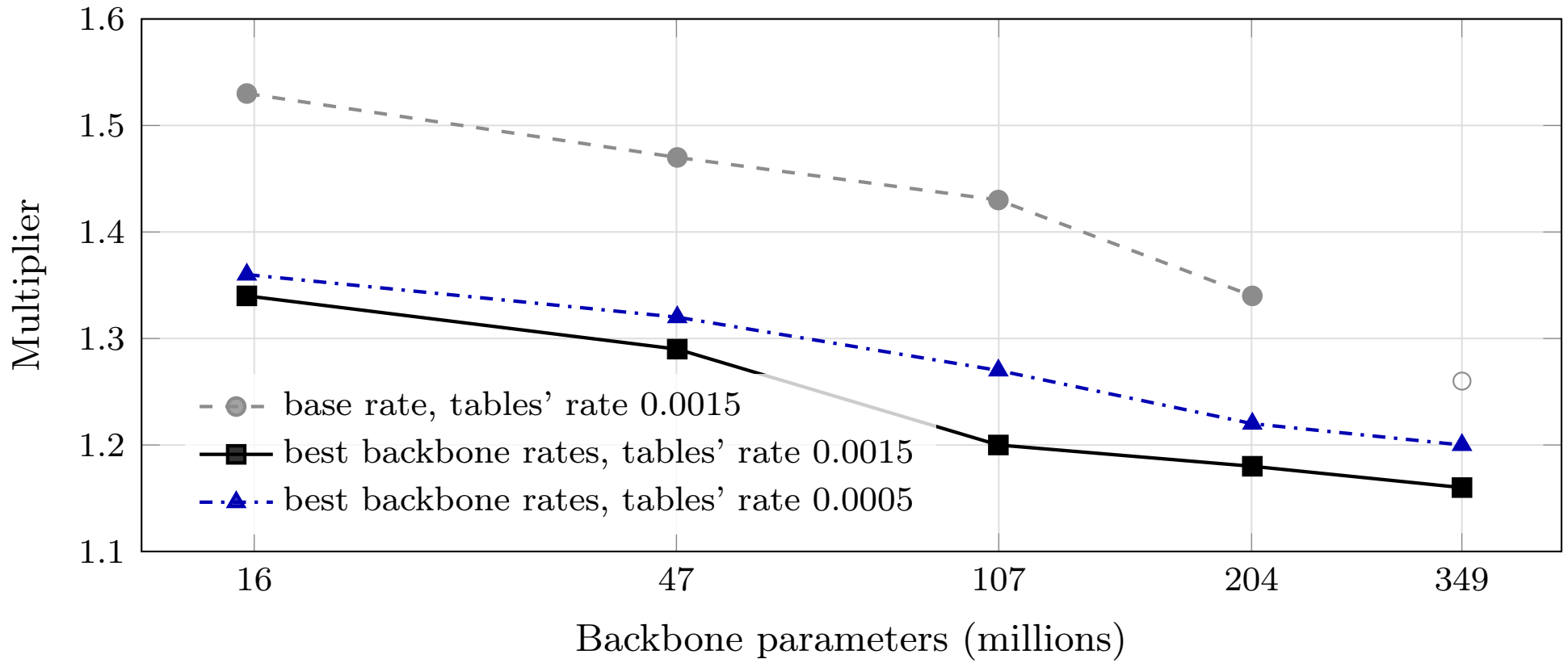}
\end{axis}
\end{tikzpicture}
\caption{Multiplier of equal-size tables against backbone size. Squares: each model at its best
measured backbone rate (Section~\ref{sec:training}), with the tables' rate of 0.0015 (Table~\ref{tab:worth}). Triangles:
the same backbone rates with a tables' rate of 0.0005, two to five seeds per size (Appendix~\ref{sec:details}).
Circles: Chorus and the plain models at the base rate; the open circle is read beyond the largest plain
model at that rate.}
\label{fig:multsize}
\end{figure}

\paragraph{Equal-size tables.} At the best measured rates, equal-size tables give a 16M model the loss
of a plain model $\times$\val{mult-16M-chorus1.09-best} its size and a 47M model that of one
$\times$\val{mult-47M-chorus1.09-best} its size. From 107M to 349M the multiplier lies between
$\times$\val{mult-349M-chorus1.09-best} and $\times$\val{mult-107M-chorus1.09-best}, lower than at the
two smaller sizes. At the base rate the multipliers are larger (lower block of Table~\ref{tab:worth}).
Figure~\ref{fig:multsize} shows these multipliers against size, with those of a lower tables' rate
(Appendix~\ref{sec:details}).
Chorus gains more on held-out windows that repeat training text word for word, and it also gains on
windows that share none (Appendix~\ref{sec:overlap}).

\paragraph{Readings.} Table~\ref{tab:readings} reads the multiplier in three ways. The first two use
the straight lines of Section~\ref{sec:analysis}, with the size counted with and without the tied
embedding. The third uses the curve $L = E + A N^{-\alpha}$, with $N$ in millions, fitted by least
squares to the six plain models at their best rates (Appendix~\ref{sec:details}). Between two measured sizes the straight lines lie above
a convex curve, so they read the larger multiplier. A change of one seed spread in one plain model moves
the multipliers next to it by a few hundredths, more than resampling the held-out windows does.

\paragraph{Prediction at 550M.} Before the 550M plain model was trained, the curve fitted to the five
smaller sizes, with the losses measured by then (Appendix~\ref{sec:details}), predicted \val{pred-550M-curve} nats for it, and the
line through 204M and 349M \val{pred-550M-line}. It reached \val{loss-550M-b-best}.

\begin{table}[H]
\centering
\small
\begin{tabular}{lrrrrr}
\toprule
 & Straight & Fitted & Without & Window & Replaced \\
Size & lines & curve & embedding & resampling (95\%) & per added \\
\midrule
16M & $\times$\val{mult-16M-chorus1.09-best} & $\times$\val{multfit-16M-chorus1.09} & $\times$\val{multne-16M-chorus1.09} & \val{ogmultci-16M} & \val{exch-16M-chorus1.09} \\
47M & $\times$\val{mult-47M-chorus1.09-best} & $\times$\val{multfit-47M-chorus1.09} & $\times$\val{multne-47M-chorus1.09} & \val{ogmultci-47M} & \val{exch-47M-chorus1.09} \\
107M & $\times$\val{mult-107M-chorus1.09-best} & $\times$\val{multfit-107M-chorus1.09} & $\times$\val{multne-107M-chorus1.09} & \val{ogmultci-107M} & \val{exch-107M-chorus1.09} \\
204M & $\times$\val{mult-204M-chorus1.09-best} & $\times$\val{multfit-204M-chorus1.09} & $\times$\val{multne-204M-chorus1.09} & \val{ogmultci-204M} & \val{exch-204M-chorus1.09} \\
349M & $\times$\val{mult-349M-chorus1.09-best} & $\times$\val{multfit-349M-chorus1.09} & $\times$\val{multne-349M-chorus1.09} & \val{ogmultci-349M} & \val{exch-349M-chorus1.09} \\
\bottomrule
\end{tabular}
\caption{Multiplier of equal-size tables at the best measured rates in three readings, and the backbone
parameters replaced per added parameter. Window resampling: 2.5th to 97.5th percentile of the straight-line
reading over 2{,}000 resamples of the held-out windows, the same for every model, without the seed spread.
Losses averaged over seeds as in Table~\ref{tab:worth} except at 16M (seeds 1337 and 2).}
\label{tab:readings}
\end{table}

\paragraph{Other budgets.} In the full runs, size and tokens per parameter change together.
Table~\ref{tab:budget} adds half and twice the budget. In each of its three pairs of sizes with about as
many tokens per parameter, the larger backbone has the lower multiplier, for instance
\val{bmult-107M-full} at 107M with the full budget against \val{bmult-47M-half} at 47M with half. Agents on
rStation report the same direction for hashed input tables, in the gain in bits per byte, with one
recipe \citeyearpar{rstation2026gp6}. At each size,
halving the budget raises the multiplier by \val{bdiff-range}, about what a plain model's seed spread
moves it. Doubling it gives \val{bmult-16M-double} at 16M and \val{bmult-47M-double} at 47M, against
\val{bmult-16M-full} and \val{bmult-47M-full}.

\begin{table}[H]
\centering
\small
\setlength{\tabcolsep}{4pt}
\begin{tabular}{lrrrrrrrrr}
\toprule
 & \multicolumn{3}{c}{Tokens per parameter} & \multicolumn{3}{c}{Gain (nats)} & \multicolumn{3}{c}{Multiplier} \\
\cmidrule(lr){2-4}\cmidrule(lr){5-7}\cmidrule(l){8-10}
Size & Half & Full & Double & Half & Full & Double & Half & Full & Double \\
\midrule
16M & 43 & 85 & 171 & 0.070 & 0.070 & 0.073 & $\times$1.38 & $\times$1.34 & $\times$1.33 \\
47M & 14 & 28 & 57 & 0.051 & 0.051 & 0.052 & $\times$1.34 & $\times$1.30 & $\times$1.28 \\
107M & 6.3 & 13 & $\cdot$ & 0.031 & 0.030 & $\cdot$ & $\times$1.25 & $\times$1.21 & $\cdot$ \\
204M & 3.3 & 6.5 & $\cdot$ & 0.026 & 0.023 & $\cdot$ & $\times$1.24 & $\times$1.19 & $\cdot$ \\
349M & 1.9 & 3.8 & $\cdot$ & 0.022 & 0.018 & $\cdot$ & $\times$1.20$^\dagger$ & $\times$1.17 & $\cdot$ \\
\bottomrule
\end{tabular}

\caption{Chorus with equal-size tables at half, the full and twice the training budget, each model at
its best measured rate at the full budget, \val{bseeds}, so the full-budget multipliers can differ from
Table~\ref{tab:readings} by a few hundredths. Each multiplier is read on the plain models of
the same budget. Tokens per parameter count the plain model's parameters. The 47M full runs saved no state at
step 2{,}297, so the 47M half runs train from step 0; at 16M a plain half run from step 0 reaches
\val{halffresh-16M} nats against \val{halfbranch-16M} for the resumed one.
$^\dagger$: read beyond the largest plain model of
that budget. $\cdot$: pair not run.}
\label{tab:budget}
\end{table}

\subsection{Larger tables}
\label{sec:tablesize}

Larger tables raise the multiplier, or leave it within a hundredth, at every size and rate where more than one table size was trained
(Figure~\ref{fig:teaser}b, Table~\ref{tab:worth}). From equal-size tables
upward, each added parameter replaces less backbone. At 47M it replaces \val{exch-47M-chorus1.09}
backbone parameters with equal-size tables and \val{exch-47M-chorusmax} with tables of 21.7. With those tables the 47M model has 1.07
billion parameters in all and still a higher loss than the 107M plain model, at the best measured
rates, \val{loss-47M-chorusmax-best} against \val{loss-107M-b-best}, and at the base rate.

\subsection{Fixed part against N-gram rows}
\label{sec:split}

The fixed-part-only rows of Table~\ref{tab:worth} split Chorus's gain between its fixed part and its
N-gram rows. At 47M and the base rate the fixed part alone gives \val{gain-47M-bve-lr1} of
Chorus's \val{gain-47M-chorus1.09-lr1} nats, about half. At Chorus's best rate it gives
\val{gain-47M-bve-best} of \val{gain-47M-chorus1.09-best}. The N-gram rows add about the same in both
settings, \val{rows-47M-lr1} nats at the base rate and \val{rows-47M-best} at the best rate, so the drop
in Chorus's gain is mostly a drop in the fixed part's gain.
With the plain model, the fixed part and Chorus all at 0.12 or 0.06 as in Section~\ref{sec:training}, the
fixed part gives \val{fixedsame-range}\% of Chorus's gain at each size where it ran.

\subsection{Vocabulary}
\label{sec:vocab}

We repeat the base-rate pairs with byte-level BPE vocabularies of 16{,}000 to 64{,}000 tokens, on the same
text and with the same tokens per step, so a larger vocabulary takes fewer steps
(Appendix~\ref{sec:details}). Chorus keeps the parameters of its 8{,}000-token
module, so its larger token table leaves fewer N-gram rows. Table~\ref{tab:vocab} gives Chorus's gain in bits
per byte, since loss per token cannot be compared between vocabularies \citep{tao2024vocab}.

\begin{table}[t]
\centering
\small
\begin{tabular}{lccccc}
\toprule
 & \multicolumn{5}{c}{Vocabulary (tokens)} \\
\cmidrule(l){2-6}
Size & 8{,}000 & 16{,}000 & 32{,}000 & 48{,}000 & 64{,}000 \\
\midrule
16M & 0.046 & 0.041 & 0.030 & $\cdot$ & $\cdot$ \\
47M & 0.033 & 0.030 & 0.025 & 0.020 & 0.018 \\
107M & 0.027 & $\cdot$ & 0.019 & $\cdot$ & 0.014 \\
349M & 0.014 & $\cdot$ & 0.013 & $\cdot$ & $\cdot$ \\
\midrule
 & \multicolumn{5}{c}{Multiplier, seed 1337} \\
\cmidrule(l){2-6}
16M & $\times$1.54 & $\times$1.45 & $\times$1.31 & $\cdot$ & $\cdot$ \\
47M & $\times$1.48 & $\times$1.30$^\dagger$ & $\times$1.34 & $\cdot$ & $\times$1.23 \\
107M & $\times$1.48 & $\cdot$ & $\times$1.33 & $\cdot$ & $\times$1.17$^\dagger$ \\
349M & $\times$1.23$^\dagger$ & $\cdot$ & $\times$1.21$^\dagger$ & $\cdot$ & $\cdot$ \\
\bottomrule
\end{tabular}

\caption{Chorus's gain over the plain model in bits per byte at the base rate (upper block). Seed 1337, averaged with seed 2 at 47M with 8{,}000 and
16{,}000 tokens. Lower block: multipliers for seed 1337, on plain models of the same vocabulary with their
embeddings counted. $^\dagger$: read beyond the largest plain model. A dot marks a pair not
trained or a multiplier not read. At 16M a token table of 48{,}000 or 64{,}000 tokens alone outgrows the module.}
\label{tab:vocab}
\end{table}

\begin{table}[t]
\centering
\small
\begin{tabular}{rrccc}
\toprule
& & \multicolumn{3}{c}{Training steps} \\
\cmidrule(l){3-5}
Vocabulary & N-gram rows & full & twice & four times \\
\midrule
8{,}000 & 74{,}820 & \val{vbudget-8k-1x} & \val{vbudget-8k-2x} & $\cdot$ \\
8{,}000 & 18{,}820 & \val{vrows-8k-few} & $\cdot$ & $\cdot$ \\
32{,}000 & 50{,}820 & \val{vbudget-32k-1x} & \val{vbudget-32k-2x} & \val{vbudget-32k-4x} \\
64{,}000 & 74{,}820 & \val{vrows-64k-many} & $\cdot$ & $\cdot$ \\
64{,}000 & 18{,}820 & \val{vbudget-64k-1x} & \val{vbudget-64k-2x} & $\cdot$ \\
\bottomrule
\end{tabular}
\caption{Chorus's gain in bits per byte at 47M, the base rate and seed 1337. N-gram rows: $m$ of
Appendix~\ref{sec:details}, 74{,}820 in the 8{,}000-token module and 18{,}820 in the 64{,}000-token one.
Longer runs start at step 0. $\cdot$: not run.}
\label{tab:vocabrows}
\end{table}

\paragraph{Gain against vocabulary.} From 16M to 107M the gain falls as the vocabulary grows, at 47M from
\val{vgain-47M-8k} with 8{,}000 tokens to \val{vgain-47M-64k} with 64{,}000. The two seeds' gains differ by up
to \val{vgain-seeddiff}. At 349M it barely changes from 8{,}000 to 32{,}000 tokens. There the 32{,}000-token
module has 8\% fewer N-gram rows, against 72\% fewer at 16M. The multipliers read within the plain models fall with the
vocabulary from 16M to 107M, and the two at 349M barely differ (lower block of Table~\ref{tab:vocab}). Table~\ref{tab:worth} differs, as it averages seeds and its line passes through 204M.

\paragraph{N-gram rows and training length.} Table~\ref{tab:vocabrows} varies the N-gram rows and the training
length at 47M. Fewer rows lower the gain at both vocabularies.
With many rows the larger vocabulary gains clearly less, and with few rows the two gains differ by about
the seed spread. Twice the steps
do not raise the gain at any of the three vocabularies, nor do four times at 32{,}000 tokens.

\subsection{Knowledge}
\label{sec:quiz}

\begin{table}[H]
\centering
\small
\setlength{\tabcolsep}{4pt}
\begin{tabular}{lrrrrrrr}
\toprule
 & \multicolumn{4}{c}{Quiz score} & \multicolumn{2}{c}{Known only by} & \\
\cmidrule(lr){2-5}\cmidrule(lr){6-7}
Size & Chorus & Plain & Equal-loss plain & Zeroed Chorus & Chorus & Plain & $p$ \\
\midrule
16M & \val{quiz-16M-chorus1.09-best} & \val{quiz-16M-b-best} & \val{quizeq-16M} & \val{quizzero-16M-chorus1.09-best} & \val{signa-16M} & \val{signb-16M} & \val{signp-16M} \\
47M & \val{quiz-47M-chorus1.09-best} & \val{quiz-47M-b-best} & \val{quizeq-47M} & \val{quizzero-47M-chorus1.09-best} & \val{signa-47M} & \val{signb-47M} & \val{signp-47M} \\
107M & \val{quiz-107M-chorus1.09-best} & \val{quiz-107M-b-best} & \val{quizeq-107M} & \val{quizzero-107M-chorus1.09-best} & \val{signa-107M} & \val{signb-107M} & \val{signp-107M} \\
204M & \val{quiz-204M-chorus1.09-best} & \val{quiz-204M-b-best} & \val{quizeq-204M} & \val{quizzero-204M-chorus1.09-best} & \val{signa-204M} & \val{signb-204M} & \val{signp-204M} \\
349M & \val{quiz-349M-chorus1.09-best} & \val{quiz-349M-b-best} & \val{quizeq-349M} & \val{quizzero-349M-chorus1.09-best} & \val{signa-349M} & \val{signb-349M} & \val{signp-349M} \\
\bottomrule
\end{tabular}
\caption{Fact quiz at the best measured rates: share of the 688 scored questions whose right answer
the model scores highest. Equal-loss plain: a plain model of the Chorus model's equivalent size, its score interpolated.
Zeroed Chorus: with its N-gram tables set to zero. Known only by: among the questions that exactly
one of the two models knows (seed 1337 of each), how many only the Chorus model knows and how many only
the plain model of its size knows. $p$: the chance of a split at least this uneven if both models were
equally good (two-sided sign test).}
\label{tab:quiz}
\end{table}

Two seeds of the 107M plain model at its best rate split \val{signseed-107M} on the sign test of
Table~\ref{tab:quiz}, so these tests also flag differences between seeds of one model. Against a plain model of their equivalent size the
differences are smaller at four of the five sizes. At 107M Chorus scores \val{quiz-107M-chorus1.09-lr2},
\val{quiz-107M-chorus1.09-lr4} and \val{quiz-107M-chorus1.09-lr8} at Muon 0.03, 0.06 and 0.12, rates
with close losses (Table~\ref{tab:lr}), so its sign test depends on which rate counts as best.

With its N-gram tables set to zero, each Chorus model scores below the plain model of its size
(Table~\ref{tab:quiz}).

\subsection{Cost}
\label{sec:cost}

A Chorus model trains \val{slowdown}\% slower than the plain model of its size, with little change
across table sizes (Table~\ref{tab:cost}).

Against a plain model of the same loss, the Chorus model with equal-size tables trains faster, by
\val{eqspeed-16M}\% at 16M and \val{eqspeed-349M}\% at 349M (Table~\ref{tab:cost}). With tables of 0.275 it trains \val{eqslow-0.275}\% slower than a plain model of
the same loss. Runs of one configuration differ in throughput by up to \val{tokspread}\%. Per
token, equal-size tables add \val{flopsadd-range}\% to the training FLOPs of the plain model of their size,
and the Chorus model needs \val{flopseq-range}\% fewer than a plain model of the same loss. Stored in four bits after training, the 47M tables change the held-out loss by
\val{quant-bits4} nats (Appendix~\ref{sec:precision}).

\begin{table}[H]
\centering
\small
\begin{tabular}{lllrrrrrr}
\toprule
 & & & \multicolumn{4}{c}{Throughput} & \multicolumn{2}{c}{FLOPs per token} \\
\cmidrule(lr){4-7}\cmidrule(l){8-9}
 & & & \multicolumn{2}{c}{tokens/s} & \multicolumn{2}{c}{Chorus relative to} & \multicolumn{2}{c}{Chorus relative to} \\
\cmidrule(lr){4-5}\cmidrule(lr){6-7}\cmidrule(l){8-9}
Size & Model & Card & Chorus & Plain & plain & equal-loss plain & plain & equal-loss plain \\
\midrule
16M & Chorus 0.275 & RTX 3080 & 250k & 288k & 0.87 &  & 1.07 & 0.98 \\
16M & Chorus 1.09 & RTX 5090 & 767k & 893k & 0.86 & 1.11 & 1.07 & 0.81 \\
16M & Chorus 4.35 & RTX 3080 & 247k & 288k & 0.86 &  & 1.07 & 0.66 \\
16M & Chorus 21.7 & RTX 5090 & 748k & 893k & 0.84 & 1.52 & 1.07 & 0.56 \\
47M & Chorus 0.275 & RTX 5090 & 304k & 339k & 0.90 & 0.95 & 1.07 & 1.00 \\
47M & Chorus 1.09 & RTX 5090 & 301k & 339k & 0.89 & 1.09 & 1.07 & 0.83 \\
47M & Chorus 4.35 & RTX 5090 & 301k & 339k & 0.89 & 1.20 & 1.07 & 0.73 \\
47M & Chorus 21.7 & RTX 5090 & 295k & 339k & 0.87 & 1.33 & 1.07 & 0.63 \\
107M & Chorus 0.275 & RTX 5090 & 159k & 179k & 0.89 & 0.91 & 1.07 & 1.05 \\
107M & Chorus 1.09 & RTX 5090 & 159k & 179k & 0.89 & 1.05 & 1.07 & 0.89 \\
107M & Chorus 4.35 & RTX 5090 & 160k & 179k & 0.89 & 1.20 & 1.07 & 0.78 \\
107M & Chorus 9.58 & RTX 5090 & 158k & 179k & 0.88 & 1.26 & 1.07 & 0.73 \\
204M & Chorus 0.275 & RTX 5090 & 90k & 100k & 0.90 & 0.94 & 1.07 & 1.03 \\
204M & Chorus 1.09 & RTX 5090 & 90k & 100k & 0.90 & 1.05 & 1.07 & 0.91 \\
204M & Chorus 4.35 & RTX 5090 & 90k & 100k & 0.91 & 1.17 & 1.07 & 0.81 \\
204M & Chorus 5.01 & RTX 5090 & 89k & 100k & 0.90 & 1.16 & 1.07 & 0.80 \\
349M & Chorus 1.09 & RTX 5090 & 56k & 62k & 0.90 & 1.03 & 1.07 & 0.92 \\
\bottomrule
\end{tabular}

\caption{Training cost of Chorus models. Throughput: tokens per second of the Chorus model and of the
plain model of its size on the same card type (medians over all runs of each size and model on that card, rates pooled). Chorus
relative to: the Chorus model's value divided by that of the plain model of its size (plain) and by
that of a plain model of the same loss (equal-loss plain), interpolated between the measured sizes;
above 1 means faster or more FLOPs. Throughput against equal-loss plain on RTX 5090 cards only.
Training FLOPs counted as in Section~\ref{sec:measure}. Blank: the equivalent size lies outside the
measured plain sizes, or no RTX 5090 run.}
\label{tab:cost}
\end{table}

\section{Summary}
\label{sec:summary}

With the backbone's rate searched for both kinds of model and the tables' rate fixed, Chorus with
equal-size tables reaches the loss of a plain model $\times$\val{mult-16M-chorus1.09-best} its size at 16M
and $\times$\val{mult-349M-chorus1.09-best} to $\times$\val{mult-107M-chorus1.09-best} its size from 107M to
349M. Each parameter added by equal-size or larger tables replaces at most about a third of a backbone
parameter, so at the
same total number of parameters a plain model reaches a lower loss, at more computation per token. With a tables' rate of 0.0005 instead of 0.0015, the multipliers are
\val{trmultup-range} higher and still fall with size. Of two backbones that see about as many tokens per parameter, the smaller at half the budget and
the larger at the full budget, the larger has the lower multiplier. Larger
tables raise the multiplier, but the 47M model with 1.07 billion parameters still has a higher loss
than the 107M plain model. A Chorus model with equal-size tables trains \val{eqspeed-range}\% faster than a plain model of the same
loss, against a run-to-run spread of up to \val{tokspread}\%. With vocabularies of 16{,}000 to 64{,}000
tokens and the module's parameters unchanged, Chorus gains less from 16M to 107M, at 47M partly because its token
table leaves fewer N-gram rows, and about as much at 349M with 32{,}000. At 47M twice the training does not raise the gain at any of the
vocabularies tried, nor does four times at 32{,}000 tokens.

\section{Limitations}
\label{sec:limits}

\begin{itemize}
\item \emph{One corpus and one module.} Every model trains on Polish printed before 1918. All but the runs
of Section~\ref{sec:vocab} use an 8{,}000-token vocabulary, and those runs use the base rate and mostly one seed.
Chorus is our own reimplementation, which departs from the authors'
code as Section~\ref{sec:chorus} describes. Its gain has the sign of the authors' and grows with table
size. We did not run the authors' setup, so we compare none of their numbers directly.
\item \emph{The rate search.} Only the backbone's Muon rate is searched, with AdamW tied to it and the
tables' rate fixed at 0.0015. At Chorus's best backbone rate, a tables' rate of 0.0005 lowered the
loss at every size, by \val{trdiff-range} nats, and raised the multiplier by \val{trmultup-range}, over
two to five seeds per size (Appendix~\ref{sec:details}). With 0.0005, the multiplier still falls with size, from
$\times$\val{trmult-16M-tr0p33} at 16M to $\times$\val{trmult-349M-tr0p33} at 349M. Part of the change
of the gain with the backbone rate may come from the fixed tables' rate. The best rate is the lowest of two to
seven losses, mostly from one seed, measured on the same held-out windows on which we report the
loss.
\item \emph{Budgets and schedule.} A full run reads 1.34 billion tokens with one short cooldown, from
85 tokens per parameter at 16M to 3.8 at 349M. The fixed part's share of what Chorus adds also changes
with backbone size. Runs at other budgets use the rates searched at the full budget.
\item \emph{Table sizes at 349M.} At 349M only equal-size tables were trained: larger tables, with two
optimiser moments per value, need more memory than a 32\,GB card has. The curve over table size
therefore ends at 204M.
\item \emph{Seeds.} Only \val{seedcells} configurations of Table~\ref{tab:runs}, at \val{seedsizes}, have more
than one seed.
\item \emph{What the subsets measure.} Windows with and without overlap also differ in their text, and
zeroing the N-gram tables also changes the input of every layer.
\item \emph{The quiz.} The author chose its facts. Facts asked in two phrasings make the sign tests
optimistic, and summing the log-probability of a multi-token answer may favour N-gram tables. Chance
is one in 47 to 72.
\item \emph{Measures.} We measure held-out loss, one quiz and training throughput on consumer cards. We
count floating-point operations from the architecture, measure no inference, and know of no standard benchmark for this
period of Polish.
\end{itemize}

\section*{Declarations}

\paragraph{Funding.} This research received no external, institutional or commercial funding. The
training compute was rented and paid for by the author or ran on the author's own machine.

\paragraph{Competing interests.} The author declares no competing interests.

\paragraph{Use of AI.} AI was used to assist with writing code and to draft and edit the text, which the
author reviewed and revised. The author directed the study, chose its research questions and
experiments, and takes full responsibility for the content.

\paragraph{Code and data.} The corpus and tokenizer are those of \citet{kocur2026wieszcz}. The code,
the configurations, the held-out results of every run and the script that computes the paper's tables
from them are at \url{https://github.com/SKocur/chorus-scaling}.

\setlength{\bibsep}{2pt plus 1pt}
\bibliographystyle{plainnat}
\bibliography{refs}

\appendix

\section{Equivalent size of every model}
\label{sec:worthall}

\begin{table}[p]
\centering
\footnotesize
\setlength{\tabcolsep}{4pt}
\setlength{\defaultaddspace}{0.2em}
\renewcommand{\arraystretch}{0.94}
\begin{tabular}{lllrrrrrrr}
\toprule
Size & Model & Muon rate & Seeds & Loss & Gain & Equivalent size & Multiplier & Quiz & Zeroed quiz \\
\midrule
\multicolumn{10}{l}{\emph{Each model at its best measured rate}} \\
\midrule
16M & plain & 0.24 & 5 & 3.5496 & & & & 0.076 & \\
16M & fixed part only & 0.12 & 1 & 3.5273 & 0.0223 & 17M & $\times$1.10 & 0.086 & $\cdot$ \\
16M & Chorus 0.275 & 0.12 & 1 & 3.5284 & 0.0212 & 17M & $\times$1.09 & 0.087 & 0.077 \\
16M & Chorus 1.09 & 0.24 & 5 & 3.4791 & 0.0705 & 21M & $\times$1.34 & 0.089 & 0.054 \\
16M & Chorus 4.35 & 0.12 & 1 & 3.4285 & 0.1211 & 26M & $\times$1.65 & 0.093 & 0.051 \\
16M & Chorus 21.7 & 0.12 & 1 & 3.3858 & 0.1638 & 31M & $\times$1.97 & 0.124 & 0.062 \\
\addlinespace
47M & plain & 0.24 & 3 & 3.2831 & & & & 0.123 & \\
47M & fixed part only & 0.12 & 1 & 3.2712 & 0.0119 & 50M & $\times$1.06 & 0.110 & $\cdot$ \\
47M & Chorus 0.275 & 0.12 & 1 & 3.2703 & 0.0128 & 50M & $\times$1.07 & 0.131 & 0.074 \\
47M & Chorus 1.09 & 0.12 & 3 & 3.2337 & 0.0494 & 61M & $\times$1.29 & 0.146 & 0.060 \\
47M & Chorus 4.35 & 0.12 & 1 & 3.2083 & 0.0748 & 70M & $\times$1.48 & 0.138 & 0.064 \\
47M & Chorus 21.7 & 0.12 & 1 & 3.1791 & 0.1039 & 81M & $\times$1.72 & 0.164 & 0.062 \\
\addlinespace
107M & plain & 0.24 & 2 & 3.1257 & & & & 0.161 & \\
107M & fixed part only & 0.06 & 1 & 3.1236 & 0.0021 & 108M & $\times$1.01 & 0.161 & $\cdot$ \\
107M & Chorus 0.275 & 0.06 & 1 & 3.1227 & 0.0030 & 109M & $\times$1.02 & 0.153 & 0.073 \\
107M & Chorus 1.09 & 0.12 & 2 & 3.0966 & 0.0291 & 129M & $\times$1.20 & 0.211 & 0.045 \\
107M & Chorus 4.35 & 0.06 & 1 & 3.0746 & 0.0511 & 148M & $\times$1.38 & 0.183 & 0.070 \\
107M & Chorus 9.58 & 0.06 & 1 & 3.0638 & 0.0619 & 158M & $\times$1.48 & 0.202 & 0.084 \\
\addlinespace
204M & plain & 0.12 & 2 & 3.0236 & & & & 0.230 & \\
204M & fixed part only & 0.06 & 1 & 3.0195 & 0.0041 & 211M & $\times$1.03 & $\cdot$ & $\cdot$ \\
204M & Chorus 0.275 & 0.06 & 1 & 3.0181 & 0.0054 & 213M & $\times$1.04 & $\cdot$ & $\cdot$ \\
204M & Chorus 1.09 & 0.06 & 2 & 3.0007 & 0.0229 & 242M & $\times$1.18 & 0.218 & 0.054 \\
204M & Chorus 4.35 & 0.06 & 1 & 2.9846 & 0.0390 & 273M & $\times$1.33 & $\cdot$ & $\cdot$ \\
204M & Chorus 5.01 & 0.06 & 1 & 2.9840 & 0.0396 & 274M & $\times$1.34 & $\cdot$ & $\cdot$ \\
\addlinespace
349M & plain & 0.12 & 2 & 2.9511 & & & & 0.233 & \\
349M & Chorus 1.09 & 0.06 & 2 & 2.9339 & 0.0172 & 404M & $\times$1.16 & 0.240 & 0.067 \\
\addlinespace
550M & plain & 0.06$^\ddagger$ & 1 & 2.8976 & & & & 0.270 & \\
\addlinespace
\midrule
\multicolumn{10}{l}{\emph{Every model at the base rate}} \\
\midrule
16M & plain & 0.015 & 1 & 3.6071 & & & & 0.071 & \\
16M & Chorus 1.09 & 0.015 & 1 & 3.4997 & 0.1074 & 24M & $\times$1.53 & 0.074 & 0.054 \\
16M & Chorus 21.7 & 0.015 & 1 & 3.3834 & 0.2236 & 38M & $\times$2.43 & 0.131 & 0.064 \\
\addlinespace
47M & plain & 0.015 & 2 & 3.3301 & & & & 0.116 & \\
47M & fixed part only & 0.015 & 2 & 3.2923 & 0.0377 & 57M & $\times$1.21 & 0.111 & $\cdot$ \\
47M & Chorus 0.275 & 0.015 & 1 & 3.2957 & 0.0343 & 56M & $\times$1.19 & 0.126 & $\cdot$ \\
47M & Chorus 1.09 & 0.015 & 2 & 3.2531 & 0.0769 & 69M & $\times$1.47 & 0.149 & 0.081 \\
47M & Chorus 4.35 & 0.015 & 1 & 3.2138 & 0.1163 & 84M & $\times$1.79 & 0.154 & 0.080 \\
47M & Chorus 21.7 & 0.015 & 2 & 3.1815 & 0.1485 & 99M & $\times$2.10 & 0.173 & 0.076 \\
\addlinespace
107M & plain & 0.015 & 1 & 3.1663 & & & & 0.135 & \\
107M & Chorus 0.275 & 0.015 & 1 & 3.1355 & 0.0308 & 127M & $\times$1.19 & 0.157 & 0.116 \\
107M & Chorus 1.09 & 0.015 & 1 & 3.1036 & 0.0627 & 153M & $\times$1.43 & 0.166 & 0.086 \\
107M & Chorus 4.35 & 0.015 & 1 & 3.0795 & 0.0868 & 175M & $\times$1.64 & 0.190 & 0.084 \\
107M & Chorus 9.58 & 0.015 & 1 & 3.0689 & 0.0974 & 186M & $\times$1.74 & 0.203 & 0.076 \\
\addlinespace
204M & plain & 0.015 & 1 & 3.0522 & & & & 0.192 & \\
204M & Chorus 1.09 & 0.015 & 1 & 3.0109 & 0.0413 & 273M & $\times$1.34 & 0.217 & 0.106 \\
\addlinespace
349M & plain & 0.015 & 1 & 2.9763 & & & & 0.227 & \\
349M & Chorus 1.09 & 0.015 & 1 & 2.9434 & 0.0329 & 440M$^\dagger$ & $\times$1.26$^\dagger$ & 0.266 & 0.109 \\
\addlinespace
\bottomrule
\end{tabular}

\caption{Equivalent size and multiplier of Chorus models and of the fixed part only, averaged over
seeds (the quiz over the seeds it was run for). Upper block: the plain model and Chorus 1.09 at their best measured rates, the other table sizes and
the fixed part only at the Muon rate given. Lower block: every model at the base rate. Chorus models are
labelled by their added ratio. Muon rate: the backbone's Muon learning rate. Gain: the plain model's
loss minus the model's loss, in nats. $^\dagger$: read beyond the largest plain model.
$^\ddagger$: best rate at the edge of the searched range. Quiz: share of the 688 scored questions
whose right answer the model scores highest. Zeroed quiz: the same model with its N-gram tables set
to zero. $\cdot$: not measured.}
\label{tab:worth}
\end{table}

\section{All runs}
\label{sec:allruns}

{\footnotesize
\setlength{\tabcolsep}{2.6pt}
\begin{longtable}{lllrrrrrrrrl}
\caption{Every run of the rate search and of the table sizes. Muon rate: the backbone's Muon learning rate at its peak (AdamW in proportion). Chorus $r$: added ratio $r$. Parameters: all of the model's, Chorus's tables included. Loss: held-out nats per token. Quiz: share of the 688 scored questions known; zeroed: with the N-gram tables set to zero. Tokens/s: training throughput (Section~\ref{sec:measure}). Micro: micro-batch in sequences, the rest of the 256 per step made up by accumulation. $\cdot$: not measured.}\label{tab:runs} \\
\toprule
Size & Model & Muon rate & Seed & Parameters & Loss & Bits/byte & Quiz & Zeroed & Tokens/s & Micro & Card \\
\midrule
\endfirsthead
\toprule
Size & Model & Muon rate & Seed & Parameters & Loss & Bits/byte & Quiz & Zeroed & Tokens/s & Micro & Card \\
\midrule
\endhead
\bottomrule
\endlastfoot
16M & plain & 0.0075 & 1337 & 15.7M & 3.6606 & 1.5760 & 0.054 & $\cdot$ & 287k & 16 & RTX 3080 \\
16M & plain & 0.015 & 1337 & 15.7M & 3.6071 & 1.5531 & 0.071 & $\cdot$ & 286k & 16 & RTX 3080 \\
16M & plain & 0.03 & 1337 & 15.7M & 3.5820 & 1.5421 & 0.064 & $\cdot$ & 288k & 16 & RTX 3080 \\
16M & plain & 0.06 & 1337 & 15.7M & 3.5685 & 1.5364 & 0.081 & $\cdot$ & 288k & 16 & RTX 3080 \\
16M & plain & 0.12 & 2 & 15.7M & 3.5595 & 1.5326 & $\cdot$ & $\cdot$ & 895k & 16 & RTX 5090 \\
16M & plain & 0.12 & 1337 & 15.7M & 3.5550 & 1.5306 & 0.077 & $\cdot$ & 894k & 16 & RTX 5090 \\
16M & plain & 0.24 & 2 & 15.7M & 3.5533 & 1.5299 & 0.076 & $\cdot$ & 893k & 16 & RTX 5090 \\
16M & plain & 0.24 & 3 & 15.7M & 3.5474 & 1.5276 & $\cdot$ & $\cdot$ & 885k & 16 & RTX 5090 \\
16M & plain & 0.24 & 4 & 15.7M & 3.5531 & 1.5299 & $\cdot$ & $\cdot$ & 885k & 16 & RTX 5090 \\
16M & plain & 0.24 & 5 & 15.7M & 3.5496 & 1.5285 & $\cdot$ & $\cdot$ & 876k & 16 & RTX 5090 \\
16M & plain & 0.24 & 1337 & 15.7M & 3.5445 & 1.5261 & 0.077 & $\cdot$ & 903k & 16 & RTX 5090 \\
16M & plain & 0.48 & 2 & 15.7M & 3.5480 & 1.5277 & $\cdot$ & $\cdot$ & 902k & 16 & RTX 5090 \\
16M & plain & 0.48 & 1337 & 15.7M & 3.5545 & 1.5305 & 0.077 & $\cdot$ & 883k & 16 & RTX 5090 \\
16M & fixed part only & 0.12 & 1337 & 20.0M & 3.5273 & 1.5187 & 0.086 & $\cdot$ & 254k & 16 & RTX 3080 \\
16M & Chorus 0.275 & 0.12 & 1337 & 20.0M & 3.5284 & 1.5193 & 0.087 & 0.077 & 250k & 16 & RTX 3080 \\
16M & Chorus 1.09 & 0.0075 & 1337 & 32.8M & 3.5280 & 1.5191 & 0.076 & 0.058 & 249k & 16 & RTX 3080 \\
16M & Chorus 1.09 & 0.015 & 1337 & 32.8M & 3.4997 & 1.5069 & 0.074 & 0.054 & 250k & 16 & RTX 3080 \\
16M & Chorus 1.09 & 0.03 & 1337 & 32.8M & 3.4892 & 1.5025 & 0.083 & 0.060 & 250k & 16 & RTX 3080 \\
16M & Chorus 1.09 & 0.06 & 2 & 32.8M & 3.4815 & 1.4991 & $\cdot$ & $\cdot$ & 769k & 16 & RTX 5090 \\
16M & Chorus 1.09 & 0.06 & 1337 & 32.8M & 3.4792 & 1.4981 & 0.080 & 0.057 & 250k & 16 & RTX 3080 \\
16M & Chorus 1.09 & 0.12 & 2 & 32.8M & 3.4804 & 1.4987 & 0.084 & 0.054 & 767k & 16 & RTX 5090 \\
16M & Chorus 1.09 & 0.12 & 1337 & 32.8M & 3.4780 & 1.4976 & 0.087 & 0.054 & 769k & 16 & RTX 5090 \\
16M & Chorus 1.09 & 0.24 & 2 & 32.8M & 3.4787 & 1.4979 & $\cdot$ & $\cdot$ & 767k & 16 & RTX 5090 \\
16M & Chorus 1.09 & 0.24 & 3 & 32.8M & 3.4783 & 1.4979 & $\cdot$ & $\cdot$ & 760k & 16 & RTX 5090 \\
16M & Chorus 1.09 & 0.24 & 4 & 32.8M & 3.4805 & 1.4988 & $\cdot$ & $\cdot$ & 764k & 16 & RTX 5090 \\
16M & Chorus 1.09 & 0.24 & 5 & 32.8M & 3.4797 & 1.4985 & $\cdot$ & $\cdot$ & 761k & 16 & RTX 5090 \\
16M & Chorus 1.09 & 0.24 & 1337 & 32.8M & 3.4781 & 1.4977 & 0.089 & 0.054 & 769k & 16 & RTX 5090 \\
16M & Chorus 1.09 & 0.48 & 1337 & 32.8M & 3.4869 & 1.5016 & 0.086 & 0.060 & 766k & 16 & RTX 5090 \\
16M & Chorus 4.35 & 0.12 & 1337 & 83.8M & 3.4285 & 1.4765 & 0.093 & 0.051 & 247k & 16 & RTX 3080 \\
16M & Chorus 21.7 & 0.015 & 1337 & 356.3M & 3.3834 & 1.4574 & 0.131 & 0.064 & 748k & 16 & RTX 5090 \\
16M & Chorus 21.7 & 0.12 & 1337 & 356.3M & 3.3858 & 1.4584 & 0.124 & 0.062 & 246k & 16 & RTX 3080 \\
47M & plain & 0.0075 & 1337 & 47.1M & 3.3633 & 1.4485 & 0.102 & $\cdot$ & 340k & 16 & RTX 5090 \\
47M & plain & 0.015 & 2 & 47.1M & 3.3283 & $\cdot$ & 0.108 & $\cdot$ & 105k & 16 & RTX 3080 \\
47M & plain & 0.015 & 1337 & 47.1M & 3.3318 & 1.4349 & 0.124 & $\cdot$ & 106k & 16 & RTX 3080 \\
47M & plain & 0.03 & 1337 & 47.1M & 3.3128 & 1.4268 & 0.106 & $\cdot$ & 339k & 16 & RTX 5090 \\
47M & plain & 0.06 & 1337 & 47.1M & 3.2946 & 1.4190 & 0.118 & $\cdot$ & 338k & 16 & RTX 5090 \\
47M & plain & 0.12 & 1337 & 47.1M & 3.2869 & 1.4158 & 0.129 & $\cdot$ & 344k & 16 & RTX 5090 \\
47M & plain & 0.24 & 2 & 47.1M & 3.2833 & 1.4142 & 0.129 & $\cdot$ & 334k & 16 & RTX 5090 \\
47M & plain & 0.24 & 3 & 47.1M & 3.2820 & 1.4136 & 0.122 & $\cdot$ & 336k & 16 & RTX 5090 \\
47M & plain & 0.24 & 1337 & 47.1M & 3.2840 & 1.4145 & 0.116 & $\cdot$ & 344k & 16 & RTX 5090 \\
47M & plain & 0.48 & 1337 & 47.1M & 3.2946 & 1.4191 & 0.128 & $\cdot$ & 338k & 16 & RTX 5090 \\
47M & fixed part only & 0.015 & 2 & 55.4M & 3.2919 & 1.4180 & 0.100 & $\cdot$ & 306k & 16 & RTX 5090 \\
47M & fixed part only & 0.015 & 1337 & 55.4M & 3.2927 & $\cdot$ & 0.122 & $\cdot$ & 96k & 16 & RTX 3080 \\
47M & fixed part only & 0.12 & 1337 & 55.4M & 3.2712 & 1.4090 & 0.110 & $\cdot$ & 308k & 16 & RTX 5090 \\
47M & fixed part only & 0.24 & 1337 & 55.4M & 3.2732 & 1.4099 & 0.112 & $\cdot$ & 307k & 16 & RTX 5090 \\
47M & Chorus 0.275 & 0.015 & 1337 & 60.1M & 3.2957 & $\cdot$ & 0.126 & $\cdot$ & 305k & 16 & RTX 5090 \\
47M & Chorus 0.275 & 0.12 & 1337 & 60.1M & 3.2703 & 1.4088 & 0.131 & 0.074 & 304k & 16 & RTX 5090 \\
47M & Chorus 1.09 & 0.0075 & 1337 & 98.5M & 3.2715 & 1.4093 & 0.129 & 0.084 & 304k & 16 & RTX 5090 \\
47M & Chorus 1.09 & 0.015 & 2 & 98.5M & 3.2530 & 1.4013 & 0.150 & $\cdot$ & 301k & 16 & RTX 5090 \\
47M & Chorus 1.09 & 0.015 & 1337 & 98.5M & 3.2532 & 1.4015 & 0.148 & 0.081 & 91k & 8 & RTX 3080 \\
47M & Chorus 1.09 & 0.03 & 1337 & 98.5M & 3.2437 & 1.3974 & 0.132 & 0.067 & 303k & 16 & RTX 5090 \\
47M & Chorus 1.09 & 0.06 & 2 & 98.5M & 3.2384 & 1.3951 & $\cdot$ & $\cdot$ & 300k & 16 & RTX 5090 \\
47M & Chorus 1.09 & 0.06 & 1337 & 98.5M & 3.2352 & 1.3937 & 0.150 & 0.054 & 302k & 16 & RTX 5090 \\
47M & Chorus 1.09 & 0.12 & 2 & 98.5M & 3.2343 & 1.3932 & 0.150 & 0.064 & 299k & 16 & RTX 5090 \\
47M & Chorus 1.09 & 0.12 & 3 & 98.5M & 3.2336 & 1.3927 & 0.151 & 0.052 & 300k & 16 & RTX 5090 \\
47M & Chorus 1.09 & 0.12 & 1337 & 98.5M & 3.2332 & 1.3929 & 0.137 & 0.062 & 303k & 16 & RTX 5090 \\
47M & Chorus 1.09 & 0.24 & 1337 & 98.5M & 3.2410 & 1.3962 & 0.138 & 0.061 & 303k & 16 & RTX 5090 \\
47M & Chorus 1.09 & 0.48 & 1337 & 98.5M & 3.2575 & 1.4033 & 0.124 & 0.058 & 300k & 16 & RTX 5090 \\
47M & Chorus 4.35 & 0.015 & 1337 & 252.1M & 3.2138 & $\cdot$ & 0.154 & 0.080 & 91k & 8 & RTX 3080 \\
47M & Chorus 4.35 & 0.12 & 1337 & 252.1M & 3.2083 & 1.3821 & 0.138 & 0.064 & 301k & 16 & RTX 5090 \\
47M & Chorus 21.7 & 0.015 & 2 & 1071.3M & 3.1808 & 1.3705 & 0.170 & 0.077 & 290k & 16 & RTX 5090 \\
47M & Chorus 21.7 & 0.015 & 1337 & 1071.3M & 3.1822 & $\cdot$ & 0.176 & 0.076 & 295k & 16 & RTX 5090 \\
47M & Chorus 21.7 & 0.12 & 1337 & 1071.3M & 3.1791 & 1.3698 & 0.164 & 0.062 & 298k & 16 & RTX 5090 \\
107M & plain & 0.015 & 1337 & 106.9M & 3.1663 & 1.3639 & 0.135 & $\cdot$ & 179k & 16 & RTX 5090 \\
107M & plain & 0.03 & 1337 & 106.9M & 3.1489 & 1.3566 & 0.147 & $\cdot$ & 178k & 16 & RTX 5090 \\
107M & plain & 0.06 & 1337 & 106.9M & 3.1378 & 1.3518 & 0.172 & $\cdot$ & 179k & 16 & RTX 5090 \\
107M & plain & 0.12 & 2 & 106.9M & 3.1247 & 1.3462 & 0.176 & $\cdot$ & 179k & 16 & RTX 5090 \\
107M & plain & 0.12 & 1337 & 106.9M & 3.1292 & 1.3480 & 0.164 & $\cdot$ & 181k & 16 & RTX 5090 \\
107M & plain & 0.24 & 2 & 106.9M & 3.1243 & 1.3460 & 0.177 & $\cdot$ & 179k & 16 & RTX 5090 \\
107M & plain & 0.24 & 1337 & 106.9M & 3.1271 & 1.3474 & 0.144 & $\cdot$ & 178k & 16 & RTX 5090 \\
107M & plain & 0.48 & 1337 & 106.9M & 3.1463 & 1.3554 & 0.170 & $\cdot$ & 178k & 16 & RTX 5090 \\
107M & fixed part only & 0.06 & 1337 & 121.1M & 3.1236 & 1.3457 & 0.161 & $\cdot$ & 162k & 16 & RTX 5090 \\
107M & Chorus 0.275 & 0.015 & 1337 & 136.3M & 3.1355 & 1.3507 & 0.157 & 0.116 & 161k & 16 & RTX 5090 \\
107M & Chorus 0.275 & 0.06 & 1337 & 136.3M & 3.1227 & 1.3454 & 0.153 & 0.073 & 158k & 16 & RTX 5090 \\
107M & Chorus 1.09 & 0.015 & 1337 & 223.4M & 3.1036 & 1.3373 & 0.166 & 0.086 & 159k & 16 & RTX 5090 \\
107M & Chorus 1.09 & 0.03 & 2 & 223.4M & 3.1033 & 1.3370 & $\cdot$ & $\cdot$ & 160k & 16 & RTX 5090 \\
107M & Chorus 1.09 & 0.03 & 1337 & 223.4M & 3.0971 & 1.3343 & 0.169 & 0.084 & 159k & 16 & RTX 5090 \\
107M & Chorus 1.09 & 0.06 & 2 & 223.4M & 3.0974 & 1.3346 & 0.190 & 0.067 & 160k & 16 & RTX 5090 \\
107M & Chorus 1.09 & 0.06 & 1337 & 223.4M & 3.0965 & 1.3340 & 0.179 & 0.061 & 160k & 16 & RTX 5090 \\
107M & Chorus 1.09 & 0.12 & 2 & 223.4M & 3.0964 & 1.3341 & $\cdot$ & $\cdot$ & 158k & 16 & RTX 5090 \\
107M & Chorus 1.09 & 0.12 & 1337 & 223.4M & 3.0968 & 1.3343 & 0.211 & 0.045 & 161k & 16 & RTX 5090 \\
107M & Chorus 1.09 & 0.24 & 1337 & 223.4M & 3.1061 & 1.3381 & $\cdot$ & $\cdot$ & 158k & 16 & RTX 5090 \\
107M & Chorus 4.35 & 0.015 & 1337 & 571.8M & 3.0795 & 1.3269 & 0.190 & 0.084 & 159k & 16 & RTX 5090 \\
107M & Chorus 4.35 & 0.06 & 1337 & 571.8M & 3.0746 & 1.3247 & 0.183 & 0.070 & 160k & 16 & RTX 5090 \\
107M & Chorus 9.58 & 0.015 & 1337 & 1131.0M & 3.0689 & 1.3223 & 0.203 & 0.076 & 157k & 16 & RTX 5090 \\
107M & Chorus 9.58 & 0.06 & 1337 & 1131.0M & 3.0638 & 1.3202 & 0.202 & 0.084 & 159k & 16 & RTX 5090 \\
204M & plain & 0.015 & 1337 & 204.4M & 3.0522 & 1.3151 & 0.192 & $\cdot$ & 100k & 8 & RTX 5090 \\
204M & plain & 0.03 & 1337 & 204.4M & 3.0364 & 1.3084 & 0.205 & $\cdot$ & 101k & 8 & RTX 5090 \\
204M & plain & 0.06 & 1337 & 204.4M & 3.0290 & 1.3051 & 0.212 & $\cdot$ & 99k & 8 & RTX 5090 \\
204M & plain & 0.12 & 2 & 204.4M & 3.0239 & 1.3030 & $\cdot$ & $\cdot$ & 99k & 8 & RTX 5090 \\
204M & plain & 0.12 & 1337 & 204.4M & 3.0232 & 1.3027 & 0.230 & $\cdot$ & 99k & 8 & RTX 5090 \\
204M & plain & 0.24 & 1337 & 204.4M & 3.0286 & 1.3050 & $\cdot$ & $\cdot$ & 100k & 8 & RTX 5090 \\
204M & fixed part only & 0.06 & 1337 & 227.1M & 3.0195 & 1.3012 & $\cdot$ & $\cdot$ & 91k & 8 & RTX 5090 \\
204M & Chorus 0.275 & 0.06 & 1337 & 260.6M & 3.0181 & 1.3006 & $\cdot$ & $\cdot$ & 90k & 8 & RTX 5090 \\
204M & Chorus 1.09 & 0.015 & 1337 & 427.2M & 3.0109 & 1.2974 & 0.217 & 0.106 & 90k & 8 & RTX 5090 \\
204M & Chorus 1.09 & 0.03 & 1337 & 427.2M & 3.0037 & 1.2943 & 0.215 & 0.084 & 90k & 8 & RTX 5090 \\
204M & Chorus 1.09 & 0.06 & 2 & 427.2M & 3.0015 & 1.2935 & $\cdot$ & $\cdot$ & 89k & 8 & RTX 5090 \\
204M & Chorus 1.09 & 0.06 & 1337 & 427.2M & 2.9999 & 1.2927 & 0.218 & 0.054 & 89k & 8 & RTX 5090 \\
204M & Chorus 1.09 & 0.12 & 1337 & 427.2M & 3.0005 & 1.2931 & 0.208 & 0.065 & 90k & 8 & RTX 5090 \\
204M & Chorus 4.35 & 0.06 & 1337 & 1093.6M & 2.9846 & 1.2862 & $\cdot$ & $\cdot$ & 90k & 8 & RTX 5090 \\
204M & Chorus 5.01 & 0.06 & 1337 & 1228.5M & 2.9840 & 1.2859 & $\cdot$ & $\cdot$ & 89k & 8 & RTX 5090 \\
349M & plain & 0.015 & 1337 & 349.1M & 2.9763 & 1.2825 & 0.227 & $\cdot$ & 62k & 8 & RTX 5090 \\
349M & plain & 0.03 & 1337 & 349.1M & 2.9617 & 1.2762 & 0.259 & $\cdot$ & 62k & 8 & RTX 5090 \\
349M & plain & 0.06 & 2 & 349.1M & 2.9552 & 1.2735 & $\cdot$ & $\cdot$ & 62k & 8 & RTX 5090 \\
349M & plain & 0.06 & 1337 & 349.1M & 2.9535 & 1.2726 & 0.243 & $\cdot$ & 62k & 8 & RTX 5090 \\
349M & plain & 0.12 & 2 & 349.1M & 2.9514 & 1.2717 & $\cdot$ & $\cdot$ & 63k & 8 & RTX 5090 \\
349M & plain & 0.12 & 1337 & 349.1M & 2.9508 & 1.2715 & 0.233 & $\cdot$ & 62k & 8 & RTX 5090 \\
349M & plain & 0.24 & 1337 & 349.1M & 2.9596 & 1.2752 & $\cdot$ & $\cdot$ & 63k & 8 & RTX 5090 \\
349M & Chorus 1.09 & 0.015 & 1337 & 729.7M & 2.9434 & 1.2685 & 0.266 & 0.109 & 56k & 8 & RTX 5090 \\
349M & Chorus 1.09 & 0.03 & 1337 & 729.7M & 2.9366 & 1.2654 & 0.275 & 0.093 & 56k & 8 & RTX 5090 \\
349M & Chorus 1.09 & 0.06 & 2 & 729.7M & 2.9352 & 1.2648 & $\cdot$ & $\cdot$ & 56k & 8 & RTX 5090 \\
349M & Chorus 1.09 & 0.06 & 1337 & 729.7M & 2.9325 & 1.2638 & 0.240 & 0.067 & 56k & 8 & RTX 5090 \\
349M & Chorus 1.09 & 0.12 & 1337 & 729.7M & 2.9348 & 1.2646 & 0.241 & 0.057 & 56k & 8 & RTX 5090 \\
550M & plain & 0.06 & 1337 & 550.4M & 2.8976 & 1.2486 & 0.270 & $\cdot$ & 41k & 8 & RTX 5090 \\
550M & plain & 0.12 & 1337 & 550.4M & 2.8987 & 1.2489 & $\cdot$ & $\cdot$ & 42k & 8 & RTX 5090 \\
\end{longtable}

\begin{longtable}{llrrrrlrr}
\caption{Every other run of the paper: other vocabularies, N-gram rows, budgets, tables' rates and Ember. N-gram rows: $m$ of Appendix~\ref{sec:details}. Steps of 256 sequences of 1{,}024 tokens. Tables' optimiser: the optimiser of the N-gram tables and its peak rate. $^\S$: continues a full run from a step at the peak rate (Table~\ref{tab:budget}). Every other run starts at step 0. Loss: held-out nats per token, comparable only within one vocabulary.}\label{tab:runsother} \\
\toprule
Size & Model & Vocab. & N-gram rows & Steps & Muon rate & Tables' optimiser & Seed & Loss \\
\midrule
\endfirsthead
\toprule
Size & Model & Vocab. & N-gram rows & Steps & Muon rate & Tables' optimiser & Seed & Loss \\
\midrule
\endhead
\bottomrule
\endlastfoot
16M & plain & 8k & & 2,552$^\S$ & 0.24 &  & 2 & 3.6377 \\
16M & plain & 8k & & 2,552$^\S$ & 0.24 &  & 1337 & 3.6271 \\
16M & plain & 8k & & 2,552 & 0.24 &  & 1337 & 3.6248 \\
16M & plain & 8k & & 10,208$^\S$ & 0.24 &  & 2 & 3.4921 \\
16M & plain & 8k & & 10,208$^\S$ & 0.24 &  & 1337 & 3.4842 \\
16M & Chorus & 8k & 33,280 & 2,552$^\S$ & 0.24 & SparseAdam 0.0015 & 2 & 3.5625 \\
16M & Chorus & 8k & 33,280 & 2,552$^\S$ & 0.24 & SparseAdam 0.0015 & 1337 & 3.5624 \\
16M & Chorus & 8k & 33,280 & 5,104 & 0.24 & SparseAdam 0.000056 & 2 & 3.4719 \\
16M & Chorus & 8k & 33,280 & 5,104 & 0.24 & SparseAdam 0.000056 & 3 & 3.4755 \\
16M & Chorus & 8k & 33,280 & 5,104 & 0.24 & SparseAdam 0.000056 & 4 & 3.4762 \\
16M & Chorus & 8k & 33,280 & 5,104 & 0.24 & SparseAdam 0.000056 & 5 & 3.4753 \\
16M & Chorus & 8k & 33,280 & 5,104 & 0.24 & SparseAdam 0.000056 & 1337 & 3.4749 \\
16M & Chorus & 8k & 33,280 & 5,104 & 0.24 & SparseAdam 0.00017 & 2 & 3.4739 \\
16M & Chorus & 8k & 33,280 & 5,104 & 0.24 & SparseAdam 0.00017 & 3 & 3.4755 \\
16M & Chorus & 8k & 33,280 & 5,104 & 0.24 & SparseAdam 0.00017 & 4 & 3.4738 \\
16M & Chorus & 8k & 33,280 & 5,104 & 0.24 & SparseAdam 0.00017 & 5 & 3.4757 \\
16M & Chorus & 8k & 33,280 & 5,104 & 0.24 & SparseAdam 0.00017 & 1337 & 3.4724 \\
16M & Chorus & 8k & 33,280 & 5,104 & 0.24 & SparseAdam 0.0005 & 2 & 3.4753 \\
16M & Chorus & 8k & 33,280 & 5,104 & 0.24 & SparseAdam 0.0005 & 3 & 3.4769 \\
16M & Chorus & 8k & 33,280 & 5,104 & 0.24 & SparseAdam 0.0005 & 4 & 3.4749 \\
16M & Chorus & 8k & 33,280 & 5,104 & 0.24 & SparseAdam 0.0005 & 5 & 3.4765 \\
16M & Chorus & 8k & 33,280 & 5,104 & 0.24 & SparseAdam 0.0005 & 1337 & 3.4740 \\
16M & Chorus & 8k & 33,280 & 10,208$^\S$ & 0.24 & SparseAdam 0.0015 & 2 & 3.4158 \\
16M & Chorus & 8k & 33,280 & 10,208$^\S$ & 0.24 & SparseAdam 0.0015 & 1337 & 3.4148 \\
16M & plain & 16k & & 4,500 & 0.015 &  & 1337 & 4.0914 \\
16M & Chorus & 16k & 25,280 & 4,500 & 0.015 & SparseAdam 0.0015 & 1337 & 3.9817 \\
16M & plain & 32k & & 4,113 & 0.015 &  & 1337 & 4.4000 \\
16M & Chorus & 32k & 9,280 & 4,113 & 0.015 & SparseAdam 0.0015 & 1337 & 4.3133 \\
47M & plain & 8k & & 2,552 & 0.24 &  & 1337 & 3.3925 \\
47M & plain & 8k & & 10,208 & 0.015 &  & 1337 & 3.2390 \\
47M & plain & 8k & & 10,208$^\S$ & 0.24 &  & 1337 & 3.2035 \\
47M & Chorus & 8k & 18,820 & 5,104 & 0.015 & SparseAdam 0.0015 & 1337 & 3.2826 \\
47M & Chorus & 8k & 74,820 & 2,552 & 0.12 & SparseAdam 0.0015 & 1337 & 3.3418 \\
47M & Chorus & 8k & 74,820 & 5,104 & 0.015 & Ember 0.0023 & 1337 & 3.2601 \\
47M & Chorus & 8k & 74,820 & 5,104 & 0.015 & Ember 0.007 & 1337 & 3.2590 \\
47M & Chorus & 8k & 74,820 & 5,104 & 0.015 & Ember 0.023 & 1337 & 3.2604 \\
47M & Chorus & 8k & 74,820 & 5,104 & 0.015 & Ember 0.073 & 1337 & 3.2626 \\
47M & Chorus & 8k & 74,820 & 5,104 & 0.015 & Ember 0.23 & 1337 & 3.2660 \\
47M & Chorus & 8k & 74,820 & 5,104 & 0.015 & Ember 0.73 & 1337 & 3.2685 \\
47M & Chorus & 8k & 74,820 & 5,104 & 0.015 & Ember lazy 0.023 & 1337 & 3.2585 \\
47M & Chorus & 8k & 74,820 & 5,104 & 0.015 & Ember lazy 0.23 & 1337 & 3.2672 \\
47M & Chorus & 8k & 74,820 & 5,104 & 0.015 & SparseAdam 0.0005 & 1337 & 3.2539 \\
47M & Chorus & 8k & 74,820 & 5,104 & 0.015 & SparseAdam 0.0045 & 2 & 3.2474 \\
47M & Chorus & 8k & 74,820 & 5,104 & 0.015 & SparseAdam 0.0045 & 1337 & 3.2522 \\
47M & Chorus & 8k & 74,820 & 5,104 & 0.015 & SparseAdam 0.0135 & 1337 & 3.2514 \\
47M & Chorus & 8k & 74,820 & 5,104 & 0.12 & SparseAdam 0.000056 & 1337 & 3.2286 \\
47M & Chorus & 8k & 74,820 & 5,104 & 0.12 & SparseAdam 0.00017 & 2 & 3.2327 \\
47M & Chorus & 8k & 74,820 & 5,104 & 0.12 & SparseAdam 0.00017 & 3 & 3.2302 \\
47M & Chorus & 8k & 74,820 & 5,104 & 0.12 & SparseAdam 0.00017 & 1337 & 3.2280 \\
47M & Chorus & 8k & 74,820 & 5,104 & 0.12 & SparseAdam 0.0005 & 2 & 3.2317 \\
47M & Chorus & 8k & 74,820 & 5,104 & 0.12 & SparseAdam 0.0005 & 3 & 3.2293 \\
47M & Chorus & 8k & 74,820 & 5,104 & 0.12 & SparseAdam 0.0005 & 1337 & 3.2290 \\
47M & Chorus & 8k & 74,820 & 5,104 & 0.12 & SparseAdam 0.0045 & 1337 & 3.2498 \\
47M & Chorus & 8k & 74,820 & 10,208 & 0.015 & SparseAdam 0.0015 & 1337 & 3.1622 \\
47M & Chorus & 8k & 74,820 & 10,208$^\S$ & 0.12 & SparseAdam 0.0015 & 1337 & 3.1513 \\
47M & plain & 16k & & 4,500 & 0.015 &  & 2 & 3.7959 \\
47M & plain & 16k & & 4,500 & 0.015 &  & 1337 & 3.7919 \\
47M & Chorus & 16k & 66,820 & 4,500 & 0.015 & SparseAdam 0.0015 & 2 & 3.7127 \\
47M & Chorus & 16k & 66,820 & 4,500 & 0.015 & SparseAdam 0.0015 & 1337 & 3.7151 \\
47M & plain & 32k & & 4,113 & 0.015 &  & 1337 & 4.1108 \\
47M & plain & 32k & & 8,226 & 0.015 &  & 1337 & 3.9766 \\
47M & plain & 32k & & 16,452 & 0.015 &  & 1337 & 3.8823 \\
47M & Chorus & 32k & 50,820 & 4,113 & 0.015 & SparseAdam 0.0015 & 1337 & 4.0376 \\
47M & Chorus & 32k & 50,820 & 8,226 & 0.015 & SparseAdam 0.0015 & 1337 & 3.9103 \\
47M & Chorus & 32k & 50,820 & 16,452 & 0.015 & SparseAdam 0.0015 & 1337 & 3.8177 \\
47M & plain & 48k & & 3,931 & 0.015 &  & 1337 & 4.2806 \\
47M & Chorus & 48k & 34,820 & 3,931 & 0.015 & SparseAdam 0.0015 & 1337 & 4.2212 \\
47M & plain & 64k & & 3,820 & 0.015 &  & 1337 & 4.3959 \\
47M & plain & 64k & & 7,640 & 0.015 &  & 1337 & 4.2451 \\
47M & Chorus & 64k & 18,820 & 3,820 & 0.015 & SparseAdam 0.0015 & 1337 & 4.3385 \\
47M & Chorus & 64k & 18,820 & 7,640 & 0.015 & SparseAdam 0.0015 & 1337 & 4.1934 \\
47M & Chorus & 64k & 74,820 & 3,820 & 0.015 & SparseAdam 0.0015 & 1337 & 4.3205 \\
107M & plain & 8k & & 2,552$^\S$ & 0.24 &  & 1337 & 3.2502 \\
107M & plain & 8k & & 10,208$^\S$ & 0.24 &  & 1337 & 3.0327 \\
107M & Chorus & 8k & 133,199 & 2,552$^\S$ & 0.12 & SparseAdam 0.0015 & 1337 & 3.2191 \\
107M & Chorus & 8k & 133,199 & 5,104 & 0.12 & SparseAdam 0.0005 & 2 & 3.0875 \\
107M & Chorus & 8k & 133,199 & 5,104 & 0.12 & SparseAdam 0.0005 & 1337 & 3.0875 \\
107M & plain & 32k & & 4,113 & 0.015 &  & 1337 & 3.9296 \\
107M & Chorus & 32k & 109,199 & 4,113 & 0.015 & SparseAdam 0.0015 & 1337 & 3.8756 \\
107M & plain & 64k & & 3,820 & 0.015 &  & 1337 & 4.2195 \\
107M & Chorus & 64k & 77,199 & 3,820 & 0.015 & SparseAdam 0.0015 & 1337 & 4.1755 \\
204M & plain & 8k & & 2,552$^\S$ & 0.12 &  & 1337 & 3.1585 \\
204M & Chorus & 8k & 208,417 & 2,552$^\S$ & 0.06 & SparseAdam 0.0015 & 1337 & 3.1324 \\
204M & Chorus & 8k & 208,417 & 5,104 & 0.06 & SparseAdam 0.0005 & 2 & 2.9975 \\
204M & Chorus & 8k & 208,417 & 5,104 & 0.06 & SparseAdam 0.0005 & 1337 & 2.9948 \\
349M & plain & 8k & & 2,552$^\S$ & 0.12 &  & 1337 & 3.0929 \\
349M & Chorus & 8k & 300,475 & 2,552$^\S$ & 0.06 & SparseAdam 0.0015 & 1337 & 3.0707 \\
349M & Chorus & 8k & 300,475 & 5,104 & 0.06 & SparseAdam 0.0005 & 2 & 2.9308 \\
349M & Chorus & 8k & 300,475 & 5,104 & 0.06 & SparseAdam 0.0005 & 1337 & 2.9289 \\
349M & plain & 32k & & 4,113 & 0.015 &  & 1337 & 3.7197 \\
349M & Chorus & 32k & 276,475 & 4,113 & 0.015 & SparseAdam 0.0015 & 1337 & 3.6827 \\
\end{longtable}

}

\section{Training details}
\label{sec:details}

\paragraph{Exact counts.} The training split has 6{,}689{,}593{,}236 tokens and the held-out split
56{,}363{,}707. Every full run reads 1{,}337{,}982{,}976 tokens: 5{,}104 steps of 256 sequences of 1{,}024
tokens, each sequence starting at a random position of the training split, in which documents are
separated by an end-of-text token.

\paragraph{Backbone.} The feed-forward layer is SwiGLU with a hidden width of 8/3 of the model width,
rounded up to a multiple of 64. Rotary embeddings use a base of 10{,}000. There is no dropout. Weights
start from a normal distribution with standard deviation 0.02, and the output projections of the
residual branches are scaled by $1/\sqrt{2L}$ for $L$ layers. RMSNorm uses $\epsilon = 10^{-5}$.

\paragraph{Optimisers.} Muon uses momentum 0.95 and five Newton--Schulz steps, and scales each update
by $\sqrt{\max(1, \text{rows}/\text{columns})}$ of the weight matrix. AdamW and SparseAdam use
$\beta = (0.9, 0.95)$. No parameter has weight decay. Gradients of every parameter except the tables
are clipped to a norm of 1.0. Training runs in \texttt{bfloat16} autocast with
\texttt{torch.compile}; the evaluation runs in \texttt{float32}. The micro-batch is 16 sequences with
16 steps of accumulation from 16M to 107M and 8 with 32 from 204M to 550M; configurations whose tables
do not fit fall back to 8 sequences with the same 256 per step.

\paragraph{Chorus.} The smallest of the four N-gram tables has $m$ rows of a quarter of the model's
width, and the others have $m+2$, $m+4$ and $m+6$. The value of $m$ is at 16M 17, 33{,}280, 166{,}334 and 875{,}943 for added
ratios of 0.275, 1.09, 4.35 and 21.7; at 47M 8{,}153, 74{,}820,
341{,}490 and 1{,}763{,}705; at 107M 19{,}771, 133{,}199, 586{,}910 and 1{,}315{,}032 (ratio 9.58); at
204M 34{,}874, 208{,}417, 902{,}595 and 1{,}043{,}140 (ratio 5.01); and at 349M 300{,}475 for equal-size tables.
At 16M and 0.275 the N-gram tables have 17 to 23 rows, so
that model is about the fixed part alone. With equal-size tables the fixed part is
\val{fixedshare-16M-chorus1.09}, \val{fixedshare-47M-chorus1.09}, \val{fixedshare-107M-chorus1.09},
\val{fixedshare-204M-chorus1.09} and \val{fixedshare-349M-chorus1.09}\% of what Chorus adds from 16M to
349M, and with tables of 0.275 at 47M it is \val{fixedshare-47M-chorus0.275}\%.

\paragraph{Fitted curve.} The curve $L = E + A N^{-\alpha}$ of Section~\ref{sec:worth} is fitted
with $\alpha$ chosen on a grid in steps of 0.002 and $E$ and $A$ by least squares for each $\alpha$,
which gives $\alpha = \val{fit-alpha}$, $E = \val{fit-E}$ and $A = \val{fit-A}$. The prediction for
550M used \val{pred-550M-inputs} nats from 16M to 349M, the lowest losses measured by then.

\paragraph{Overlap.} Checked token by token, sampled hash matches between held-out and training
sequences agree in \val{ov13-checked} cases for 13 tokens and \val{ov50-checked} for 50.

\paragraph{Cards.} Runs were trained on RTX 5090 cards (32\,GB) and on one RTX 3080 (10\,GB);
Table~\ref{tab:runs} lists the card of every run.

\paragraph{Bits per byte.} The bits per byte of Table~\ref{tab:runs} cover every window of the held-out
split, more text than the 8{,}192 windows of the loss: the summed cross-entropy divided by $\ln 2$ and by
the UTF-8 bytes of the predicted tokens.

\paragraph{Vocabularies.} The vocabularies of 16{,}000 to 64{,}000 tokens are byte-level BPE, trained with
one recipe on a 4\,GB sample of a later version of the corpus, cleaned and with more sources. The
8{,}000-token tokenizer of \citet{kocur2026wieszcz} was trained on a sample of the earlier version. The sample
was drawn without excluding the held-out documents, which affects both arms alike. The
training and held-out splits are the 8{,}000-token streams decoded and encoded again, so every
vocabulary sees the same text. Steps and warmup shrink by the ratio of tokens: 4{,}500, 4{,}113,
3{,}931 and 3{,}820 steps for 16{,}000 to 64{,}000 tokens. Chorus loses one N-gram row for each added
token: at 47M its 74{,}820 rows become 66{,}820, 50{,}820, 34{,}820 and 18{,}820. The gains of
Section~\ref{sec:vocab} are differences of held-out loss in nats, divided by $\ln 2$ and by the
vocabulary's bytes per token on the held-out split.

\paragraph{The tables' rate.} At 47M and the base rate, Chorus with equal-size tables reached
\val{side-tablerate1} nats with the tables' rate of 0.0015 (two seeds), and \val{side-tablerate0p33},
\val{side-tablerate3} and \val{side-tablerate9} nats with 0.0005, 0.0045 and 0.0135 (seed 1337, and seed 2 at 0.0045). At a Muon rate of
0.12, Chorus's best at 47M, tables' rates of 0.0005 and 0.0045 gave \val{side-tablerate0p33-lr8} and
\val{side-tablerate3-lr8} nats against \val{side-tablerate1-lr8} at 0.0015, all for seed 1337.
At every size, with equal-size tables and Chorus's best backbone rate, lower tables' rates gave:

\begin{center}
\small
\setlength{\tabcolsep}{4pt}
\begin{tabular}{llcrrrrrr}
\toprule
 & & & \multicolumn{2}{c}{Chorus loss (nats)} & \multicolumn{2}{c}{Equivalent size} & \multicolumn{2}{c}{Multiplier} \\
\cmidrule(lr){4-5}\cmidrule(lr){6-7}\cmidrule(l){8-9}
Size & Tables' rate & Seeds & at 0.0015 & lower & at 0.0015 & lower & at 0.0015 & lower \\
\midrule
16M & 0.0005 & \val{trseeds-16M-tr0p33} & \val{trlossdef-16M-tr0p33} & \val{trloss-16M-tr0p33} & \val{treqdef-16M-tr0p33} & \val{treq-16M-tr0p33} & $\times$\val{trmultdef-16M-tr0p33} & $\times$\val{trmult-16M-tr0p33} \\
16M & 0.00017 & \val{trseeds-16M-tr0p11} & \val{trlossdef-16M-tr0p11} & \val{trloss-16M-tr0p11} & \val{treqdef-16M-tr0p11} & \val{treq-16M-tr0p11} & $\times$\val{trmultdef-16M-tr0p11} & $\times$\val{trmult-16M-tr0p11} \\
16M & 0.000056 & \val{trseeds-16M-tr0p04} & \val{trlossdef-16M-tr0p04} & \val{trloss-16M-tr0p04} & \val{treqdef-16M-tr0p04} & \val{treq-16M-tr0p04} & $\times$\val{trmultdef-16M-tr0p04} & $\times$\val{trmult-16M-tr0p04} \\
47M & 0.0005 & \val{trseeds-47M-tr0p33} & \val{trlossdef-47M-tr0p33} & \val{trloss-47M-tr0p33} & \val{treqdef-47M-tr0p33} & \val{treq-47M-tr0p33} & $\times$\val{trmultdef-47M-tr0p33} & $\times$\val{trmult-47M-tr0p33} \\
47M & 0.00017 & \val{trseeds-47M-tr0p11} & \val{trlossdef-47M-tr0p11} & \val{trloss-47M-tr0p11} & \val{treqdef-47M-tr0p11} & \val{treq-47M-tr0p11} & $\times$\val{trmultdef-47M-tr0p11} & $\times$\val{trmult-47M-tr0p11} \\
107M & 0.0005 & \val{trseeds-107M-tr0p33} & \val{trlossdef-107M-tr0p33} & \val{trloss-107M-tr0p33} & \val{treqdef-107M-tr0p33} & \val{treq-107M-tr0p33} & $\times$\val{trmultdef-107M-tr0p33} & $\times$\val{trmult-107M-tr0p33} \\
204M & 0.0005 & \val{trseeds-204M-tr0p33} & \val{trlossdef-204M-tr0p33} & \val{trloss-204M-tr0p33} & \val{treqdef-204M-tr0p33} & \val{treq-204M-tr0p33} & $\times$\val{trmultdef-204M-tr0p33} & $\times$\val{trmult-204M-tr0p33} \\
349M & 0.0005 & \val{trseeds-349M-tr0p33} & \val{trlossdef-349M-tr0p33} & \val{trloss-349M-tr0p33} & \val{treqdef-349M-tr0p33} & \val{treq-349M-tr0p33} & $\times$\val{trmultdef-349M-tr0p33} & $\times$\val{trmult-349M-tr0p33} \\
\bottomrule
\end{tabular}
\end{center}

\noindent ``Lower'' is the tables' rate of the row. The equivalent size is that of a plain model with the
same loss on the line through the plain models at their best rates, and the multiplier divides it by
the backbone's size. Both columns of a pair come from the same seeds.

\paragraph{Ember.} We trained Chorus with equal-size tables at 47M and the base rate with Ember at six
rates, at seed 1337. Every rate gave a higher held-out loss than SparseAdam at 0.0015, which reached
\val{side-tablerate1} nats. The rate of 0.23 is the authors' $0.2\,(d/768)^{-1/2}$ at our width
$d = 576$. In the authors' form every row's second-moment value decays at each step, whether the row is
read or not. In the lazy form, which we added, it changes only when the row is read, and its bias
correction counts that row's reads.

\begin{center}
\small
\begin{tabular}{lcccccc}
\toprule
Ember rate & 0.0023 & 0.007 & 0.023 & 0.073 & 0.23 & 0.73 \\
\midrule
Authors' form & \val{side-ember-lr0p0023} & \val{side-ember-lr0p007} & \val{side-ember-lr0p023}
 & \val{side-ember-lr0p073} & \val{side-ember} & \val{side-ember-lr0p73} \\
Lazy form & $\cdot$ & $\cdot$ & \val{side-emberlazy-lr0p023} & $\cdot$ & \val{side-emberlazy} & $\cdot$ \\
\bottomrule
\end{tabular}
\end{center}

\noindent Held-out loss in nats; a dot marks a rate not run in that form.

\section{Tables at lower precision}
\label{sec:precision}

We round every row of the tables, the token table included, to signed integers of a given width, with
the row's largest magnitude as its 16-bit scale, and measure the change in held-out loss on 1{,}024
windows spread over the held-out split (Table~\ref{tab:quant}). Stored in eight bits, the tables
change the loss by \val{quant-bits8} nats. In four bits the change is \val{quant-bits4} nats, at about
a quarter of the tables' 16-bit size and with the number of right quiz answers changed by at most \val{quantquiz-bits4}; the tables keep
\val{quantkept-bits4}\% of the gain of the same configuration on all windows. In three bits the change
is \val{quant-bits3} nats, and in two bits \val{quant-bits2}, more than the tables gave.
\citet{chung2026lowbit} report more than 99\% of the gain of generic N-gram tables kept at four bits,
as a share of the tables' own gain; ours is a share of the whole module's gain, at the base rate.

\begin{table}[H]
\centering
\small
\begin{tabular}{llrrr}
\toprule
Table & Bits & Loss change & Quiz & Table size \\
\midrule
47M, Chorus 1.09 & 32 (trained) &  & 0.148 &  \\
47M, Chorus 1.09 & 8 & +0.0000 & 0.148 & 48 MB (bf16 95 MB) \\
47M, Chorus 1.09 & 4 & +0.0021 & 0.148 & 25 MB (bf16 95 MB) \\
47M, Chorus 1.09 & 3 & +0.0122 & 0.144 & 19 MB (bf16 95 MB) \\
47M, Chorus 1.09 & 2 & +0.1864 & 0.103 & 13 MB (bf16 95 MB) \\
\addlinespace
47M, Chorus 4.35 & 32 (trained) &  & 0.154 &  \\
47M, Chorus 4.35 & 8 & +0.0000 & 0.154 & 204 MB (bf16 403 MB) \\
47M, Chorus 4.35 & 4 & +0.0022 & 0.153 & 103 MB (bf16 403 MB) \\
47M, Chorus 4.35 & 3 & +0.0128 & 0.156 & 78 MB (bf16 403 MB) \\
47M, Chorus 4.35 & 2 & +0.2152 & 0.125 & 53 MB (bf16 403 MB) \\
\addlinespace
47M, Chorus 21.7 & 32 (trained) &  & 0.170 &  \\
47M, Chorus 21.7 & 8 & +0.0000 & 0.170 & 1035 MB (bf16 2041 MB) \\
47M, Chorus 21.7 & 4 & +0.0025 & 0.172 & 524 MB (bf16 2041 MB) \\
47M, Chorus 21.7 & 3 & +0.0141 & 0.170 & 397 MB (bf16 2041 MB) \\
47M, Chorus 21.7 & 2 & +0.2351 & 0.131 & 269 MB (bf16 2041 MB) \\
\addlinespace
\bottomrule
\end{tabular}

\caption{Chorus tables of three 47M models trained at the base rate (added ratios 1.09 and 4.35 from
seed 1337, 21.7 from seed 2), stored at lower precision after training: change in held-out loss on 1{,}024 windows spread over the
held-out split against the trained tables, quiz score and table size.}
\label{tab:quant}
\end{table}

\section{Overlap between held-out and training text}
\label{sec:overlap}

An N-gram memory can gain most where the held-out text repeats the training text word for
word, as in reprints. For each of the 8{,}192 scored windows we therefore find which of its 13-token and 50-token
sequences occur anywhere in the training split (Appendix~\ref{sec:details}). A run reads
about a fifth of the split, so the shares below bound the overlap with what any one run saw. Of the
13-token sequences at all positions of the windows, \val{ov13-share}\% occur in the training split, and \val{ov13-over50} windows have
more than half of theirs there. Of the 50-token sequences \val{ov50-share}\% occur there, in
\val{ov50-any} windows.

Chorus gains more on windows that share text with the training split, and it also gains on windows that
share none (Table~\ref{tab:overlap}). At 47M its gain falls from \val{oggain-47M-all} nats over all windows
to \val{oggain-47M-50gramnone} on the \val{ogwin-50gramnone} windows with no shared 50-token sequence,
and to \val{oggain-47M-13gramnone} on the \val{ogwin-13gramnone} with no shared 13-token sequence. Read on the
windows without a shared 50-token sequence, each multiplier of equal-size tables is a little lower than
over all windows.

\begin{table}[H]
\centering
\small
\begin{tabular}{lrrrrrr}
\toprule
 & \multicolumn{3}{c}{Gain (nats)} & \multicolumn{3}{c}{Multiplier} \\
\cmidrule(lr){2-4}\cmidrule(l){5-7}
Size & All windows & No shared 50 & No shared 13 & All windows & No shared 50 & No shared 13 \\
\midrule
16M & \val{oggain-16M-all} & \val{oggain-16M-50gramnone} & \val{oggain-16M-13gramnone} & $\times$\val{ogmult-16M} & $\times$\val{ogmult-16M-50gramnone} & $\times$\val{ogmult-16M-13gramnone} \\
47M & \val{oggain-47M-all} & \val{oggain-47M-50gramnone} & \val{oggain-47M-13gramnone} & $\times$\val{ogmult-47M} & $\times$\val{ogmult-47M-50gramnone} & $\times$\val{ogmult-47M-13gramnone} \\
107M & \val{oggain-107M-all} & \val{oggain-107M-50gramnone} & \val{oggain-107M-13gramnone} & $\times$\val{ogmult-107M} & $\times$\val{ogmult-107M-50gramnone} & $\times$\val{ogmult-107M-13gramnone} \\
204M & \val{oggain-204M-all} & \val{oggain-204M-50gramnone} & \val{oggain-204M-13gramnone} & $\times$\val{ogmult-204M} & $\times$\val{ogmult-204M-50gramnone} & $\times$\val{ogmult-204M-13gramnone} \\
349M & \val{oggain-349M-all} & \val{oggain-349M-50gramnone} & \val{oggain-349M-13gramnone} & $\times$\val{ogmult-349M} & $\times$\val{ogmult-349M-50gramnone} & $\times$\val{ogmult-349M-13gramnone} \\
\midrule
Windows & \val{ogwin-all} & \val{ogwin-50gramnone} & \val{ogwin-13gramnone} & & & \\
\bottomrule
\end{tabular}
\caption{Chorus's gain with equal-size tables and its multiplier at the best measured rates, over all
8{,}192 held-out windows and over the windows that share no 50-token or no 13-token sequence with the
training split; the plain models are read on the same windows, losses averaged over seeds as in
Table~\ref{tab:worth} except at 16M (seeds 1337 and 2).}
\label{tab:overlap}
\end{table}

\section{The fact quiz}
\label{sec:quizitems}

Table~\ref{tab:quizitems} gives one question of each group. The model reads the Polish opening
and scores each candidate as its continuation; the English line is a translation for the reader.
Facts of the capitals, rivers and authors groups are also asked in a second phrasing, for example
\emph{Głównem miastem Francyi jest} (``The main city of France is'').

\begin{table}[H]
\centering
\small
\begin{tabular}{p{0.17\linewidth}p{0.47\linewidth}p{0.16\linewidth}r}
\toprule
Group & Opening (translation) & Answer & Candidates \\
\midrule
Capitals & \emph{Stolicą Francyi jest} (The capital of France is) & Paryż (Paris) & 54 \\
Rivers & \emph{Kraków leży nad} (Kraków lies on the) & Wisłą (Vistula) & 48 \\
Authors & \emph{Autor dzieła „Pan Tadeusz'' zwał się} (The author of the work Pan Tadeusz was called) & Mickiewicz & 72 \\
Dates & \emph{Jan Sobieski pobił Turków pod Wiedniem w roku} (Jan Sobieski defeated the Turks at Vienna in the year) & 1683 & 70 \\
Persons & \emph{Ostatnim królem polskim był Stanisław August} (The last Polish king was Stanisław August) & Poniatowski & 49 \\
Scripture and antiquity & \emph{Patryarcha, który zbudował arkę przed potopem, zwał się} (The patriarch who built the ark before the flood was called) & Noe (Noah) & 47 \\
Nature and everyday & \emph{Ile nóg ma pająk? Pająk ma} (How many legs has a spider? A spider has) & ośm (eight) & none \\
\bottomrule
\end{tabular}
\caption{One question from each group of the fact quiz, in the spelling of the period (\emph{Francyi},
\emph{Patryarcha}, \emph{ośm}). Candidates: the answers of the group, all scored as continuations of the
opening. The nature and everyday group has no candidate list and is left out of the score.}
\label{tab:quizitems}
\end{table}

\section{Further related work}
\label{sec:further}

\begin{table}[H]
\centering
\small
\begin{tabular}{>{\raggedright\arraybackslash}p{0.3\linewidth}>{\raggedright\arraybackslash}p{0.62\linewidth}}
\toprule
Work & What it does \\
\midrule
SCONE \citep{yu2025scaling} & embeddings of frequent N-grams computed by a separate model and cached \\
Byte Latent Transformer \citep{pagnoni2025byte} & hashed N-gram embeddings added to byte embeddings \\
N-gram embeddings against experts \citep{liu2026scaling} & N-gram tables compared with more experts at a fixed total \\
Tensorizing Engram (TN-gram) \citep{zhou2026tensorizing}, FactorEngram \citep{yang2026factorengram} & factorised N-gram tables \\
Low-bit N-gram embeddings \citep{chung2026lowbit} & N-gram tables stored in four bits \\
Product keys \citep{lample2019large}, memory layers \citep{berges2025memory}, UltraMem \citep{huang2025ultrasparse} & large sparse memories read by learned keys \\
TIDE \citep{jaiswal2026tide} & token-indexed memories injected into every layer \\
Infini-gram \citep{liu2024infinigram} & count-based N-gram model of unbounded order \\
\bottomrule
\end{tabular}
\caption{Further work on token-indexed and N-gram memories.}
\end{table}

\end{document}